\documentclass{article}

\PassOptionsToPackage{numbers}{natbib}
\usepackage[preprint]{neurips_2026}

\usepackage{microtype}
\usepackage{graphicx}
\usepackage{subcaption}
\usepackage{booktabs}
\usepackage{multirow}
\usepackage{array}
\usepackage{amsmath}
\usepackage{amssymb}
\usepackage{mathtools}
\usepackage{amsthm}
\usepackage{algorithm}
\usepackage{algorithmic}
\usepackage{float}
\usepackage{wrapfig}
\usepackage{needspace}
\usepackage{relsize}
\usepackage{longtable}
\usepackage{xcolor}
\usepackage{hyperref}
\usepackage{url}
\usepackage[capitalize,noabbrev]{cleveref}
\usepackage{pgfplots}
\pgfplotsset{compat=1.17}
\usepackage{colortbl}

\definecolor{skipColorSR}{HTML}{2E7D32}    
\definecolor{skipColorSteps}{HTML}{1565C0} 

\newcolumntype{C}[1]{>{\centering\arraybackslash}p{#1}}
\newcommand{\Task}[1]{\texttt{#1}}

\DeclareRobustCommand{\TaskUS}[1]{{\ttfamily\def\_{\char`\_\allowbreak}#1}}
\newcommand{\cmark}{\ensuremath{\checkmark}}
\newcommand{\ms}[2]{#1{\smaller\,$\pm$\,#2}}

\theoremstyle{plain}

\theoremstyle{definition}

\theoremstyle{remark}

\title{SkiP: When to Skip and When to Refine\\for Efficient Robot Manipulation}

\author{
  \textbf{Mingtong Dai\textsuperscript{1,2,6}}
  \quad
  \textbf{Guanqi Peng\textsuperscript{3}}
  \quad
  \textbf{Yongjie Bai\textsuperscript{2,4}}
  \quad
  \textbf{Feng Yan\textsuperscript{5}}
  \\
  \textbf{Chunjie Chen\textsuperscript{1}}
  \quad
  \textbf{Lingbo Liu\textsuperscript{2}}\thanks{Corresponding author.}
  \quad
  \textbf{Liang Lin\textsuperscript{2,4}}
  \quad
  \textbf{Xinyu Wu\textsuperscript{1}}
  \\
  \textsuperscript{1}Shenzhen Institutes of Advanced Technology, Chinese Academy of Sciences
  \\
  \textsuperscript{2}Peng Cheng Laboratory
  \quad
  \textsuperscript{3}Southern University of Science and Technology
  \\
  \textsuperscript{4}Sun Yat-sen University
  \quad
  \textsuperscript{5}UNT
  \\
  \textsuperscript{6}University of Chinese Academy of Sciences
}

\begin{document}

\maketitle

\begin{abstract}
Previous imitation learning policies predict future actions at every control step, whether in smooth motion phases or precise, contact-rich operation phases.
This uniform treatment is wasteful: most steps in a manipulation trajectory traverse free space and carry little task-relevant information, while a small fraction of \emph{key} steps around contacts, grasps, and alignment demand dense, high-resolution prediction.
We propose a novel \emph{action relabeling} mechanism: at each timestep in a skip segment, we replace the behavior cloning target with the action at the entrance of the next key segment, enabling the policy to leap over redundant steps in a single decision.
The resulting \textbf{Skip Policy (SkiP)} dynamically leaps over skip segments and intensively refines actions in key segments, within a single unified network requiring no learned skip planner or hierarchical structure.
To automatically partition demonstrations into key and skip segments without manual annotation, we introduce \emph{Motion Spectrum Keying} (MSK), a fast, task-agnostic procedure that detects local motion complexity from action signals.
Extensive experiments across 72 simulated manipulation tasks and three real-robot tasks show that SkiP reduces executed steps by $15$--$40\%$ while matching or improving success rates across various policy backbones. Project page: \texttt{https://pgq18.github.io/SkiP-page/}.
\end{abstract}

\section{Introduction}
\label{sec:intro}

A robot reaching for a bottle on a shelf traverses half a meter of empty air in a smooth arc, then must carefully close its gripper around the neck.
Transit through free space and contact-rich manipulation demand fundamentally different control resolutions, yet behavior cloning treats every timestep identically: one observation in, one action out, repeated hundreds of times per episode.
In free-space segments, step-by-step prediction is redundant and each additional policy query compounds prediction error~\citep{Efficient-Reductions-for-Imitation-Learning,A-Reduction-of-IL}.
In contact segments, the policy must react densely to maintain precision.
Standard imitation learning satisfies neither need well.

How should a policy allocate its decisions across these regimes?
Prior work answers by enriching the \emph{model}: action chunking predicts short action windows~\citep{zhao2023act,chi2023diffusion,zhang2024arp}; keyframe methods factor trajectories into sparse anchors and dense connectors~\citep{xian2023chaineddiffuser,zhang2025coa}; hierarchical policies add planners or options on top of low-level controllers~\citep{sutton1999between}.
These approaches improve capacity, but the underlying policy still \emph{executes} at a uniform temporal rate: every timestep receives the same treatment regardless of its information content.
Speed-adaptive methods~\citep{arachchige2025sail} adjust velocity but do not modify what the policy learns to predict.
In short, prior methods change how the policy generates actions, not \emph{what} it is supervised to predict.

\begin{figure}[t]
\centering
\includegraphics[width=0.88\columnwidth]{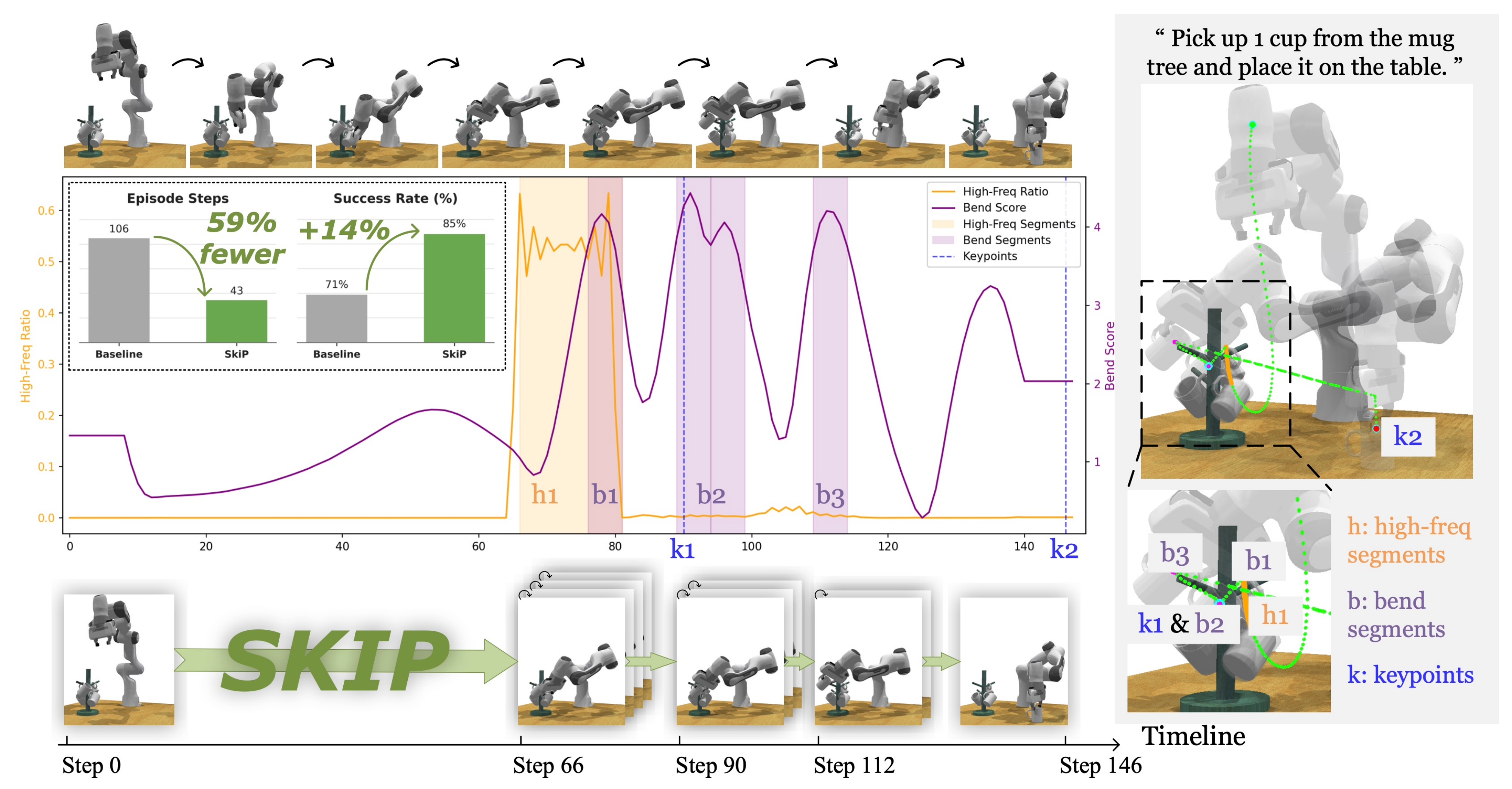}
\vspace{-1ex}
\caption{``Pick up 1 cup from the mug tree and place it on the table'' analyzed by SkiP. \textbf{Top:} high-frequency energy ratio (orange) and bend score (purple); shaded regions are key segments. \textbf{Bottom:} SkiP skips from step 0 to step 66, then refines through contacts.}
\label{fig:teaser}
\end{figure}

\begin{figure}[t]
\centering
\includegraphics[width=\columnwidth]{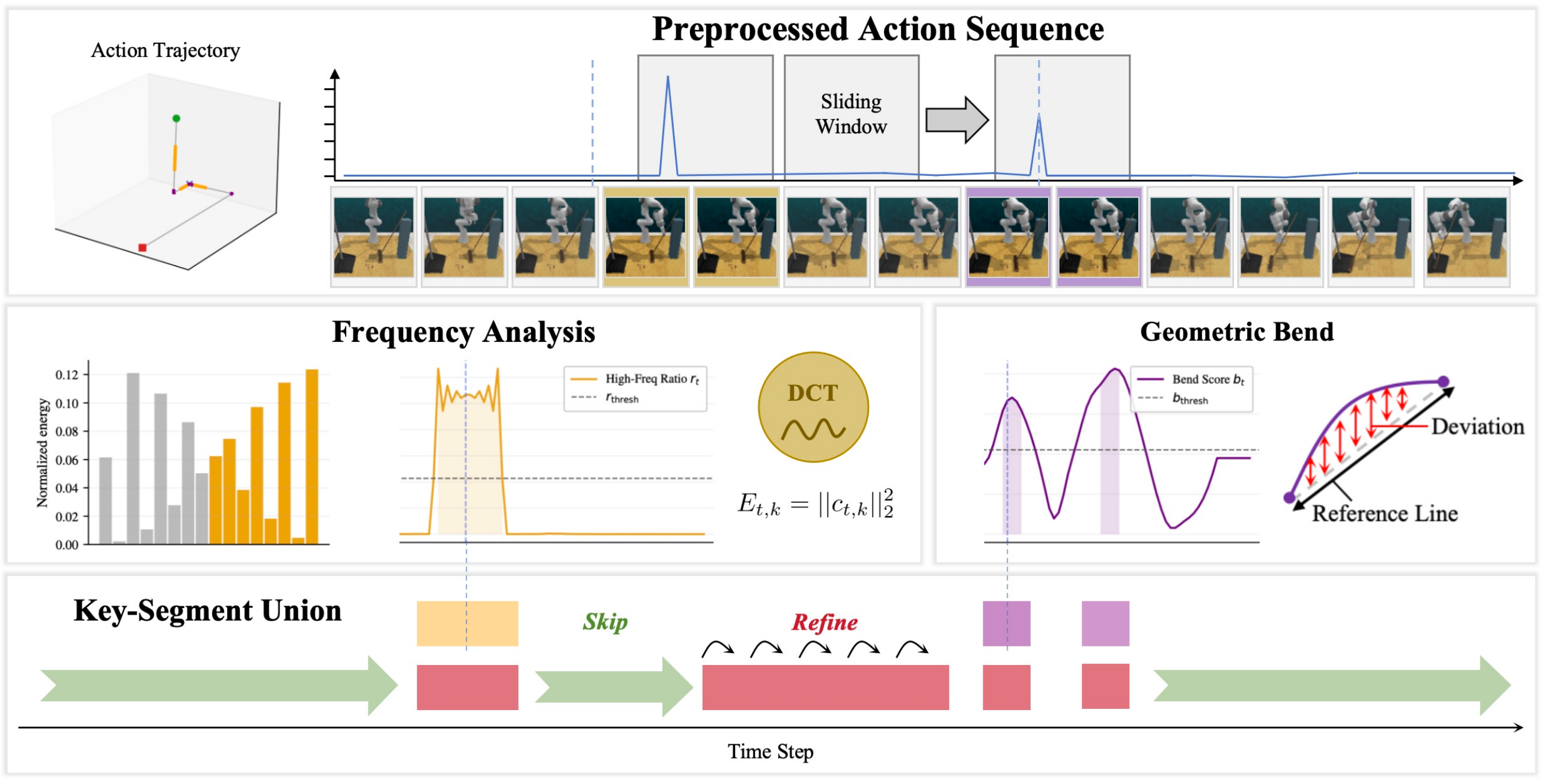}
\caption{Overview of SkiP. We partition each demonstration into high-information \emph{key segments} and low-information \emph{skip segments} via DCT spectral analysis, then \emph{relabel} training targets: in skip segments, the target is the action at the next key-segment entrance; in key segments, the target is the next-step continuation.}
\label{fig:pipeline}
\end{figure}

We take a different route (Figures~\ref{fig:teaser} and~\ref{fig:pipeline}).
Rather than adding architectural complexity, we modify the \emph{training target} itself.
At each timestep in a skip segment, we replace the behavior cloning target with the action at the entrance of the next key segment; inside key segments, the target remains the immediate next step.
We call this \emph{action relabeling}.
The resulting policy, \textbf{SkiP} (Skip Policy), learns \emph{when to skip}, advancing past predictable motion in a single decision, and \emph{when to refine}, predicting dense corrections near contacts.
It does so within a single network, requiring no architectural changes, no learned skip planner, and no additional inference cost.

Skipping also \emph{improves accuracy}: each policy query is an opportunity for prediction error to accumulate, and collapsing a free-space traverse into one decision removes the intermediate predictions that would otherwise compound.
This explains why SkiP often \emph{raises} success rate while cutting executed steps.

To partition demonstrations into key and skip segments, we introduce \emph{Motion Spectrum Keying} (MSK), which identifies key segments by measuring local frequency content of the action signal.
MSK is fast, task-agnostic, and requires no learned components.

Our contributions are as follows:
\begin{enumerate}
\item \textbf{Action relabeling.} We show that modifying the behavior cloning target alone is sufficient to produce adaptive skip-and-refine behavior, without architectural changes or learned planners. This demonstrates that temporal resolution in imitation learning can be controlled through the training objective.
\item \textbf{Motion Spectrum Keying (MSK).} A frequency-domain procedure that partitions demonstrations into key and skip segments without learning, connecting spectral signal processing to imitation learning supervision design.
\item \textbf{Strong empirical performance.} SkiP achieves state-of-the-art results across extensive robot manipulation benchmarks, improving both success rate and execution efficiency while preserving the same policy backbones. We further analyze the learned action magnitudes and show that the policy acquires distinct skip and refine modes (\S\ref{sec:exp:analysis}).
\end{enumerate}

\section{Related Work}
\label{sec:related}

\paragraph{Dense visuomotor imitation.}
Transformer-based manipulation policies such as PerAct~\citep{shridhar2022peract}, HiveFormer~\citep{guhur2022hiveformer}, and Act3D~\citep{gervet2023act3d} predict 6-DoF actions from structured 3D representations or language-conditioned histories.
RVT~\citep{goyal2023rvt} and TVVE~\citep{bai2025tvve} further improve view-based 3D manipulation by predicting keyframe-style actions from compact visual representations.
Language-conditioned manipulation systems~\citep{shridhar2022cliport,liu2024robouniview,yan2025robotronmani} combine semantic task understanding with spatially precise action prediction.
Diffusion Policy~\citep{chi2023diffusion}, 3D Diffuser Actor~\citep{ke2024diffuseractor}, and Action Chunking with Transformers (ACT)~\citep{zhao2023act} model short action windows via denoising or autoregressive prediction; this reduces replanning frequency but execution remains uniform.
Behavior Transformer~\citep{shafiullah2022behavior} and other autoregressive sequence models further explore multimodal or variable-length action prediction~\citep{zhang2024arp}.
At a larger scale, complex manipulation benchmarks~\citep{rlbench,mees2022calvin,mandlekar2021robomimic,mu2024robotwin,zheng2024robocas} and generalist policies~\citep{brohan2022rt1,brohan2023rt2,octo2024octo,kim2024openvla} show that broad data and vision-language pretraining improve visuomotor control.
These works mainly change the action generator, visual representation, or data scale, whereas SkiP changes the supervision target used to allocate temporal resolution.

\paragraph{Keyframe trajectory structuring and temporal abstraction.}
Temporal abstraction through sparse anchors is a classical strategy for long rollouts; in reinforcement learning it is formalized through the options framework~\citep{sutton1999between}.
ChainedDiffuser~\citep{xian2023chaineddiffuser} unifies keypose prediction with a diffusion module that fills in connecting segments.
Chain-of-Action (CoA)~\citep{zhang2025coa} generates full trajectories via backward reasoning from a predicted keyframe under a trajectory autoregressive formulation.
Keyframe-focused imitation~\citep{karaev2021keyframe} and automated annotation pipelines~\citep{zuo2024kisa} emphasize key moments in demonstrations.
These methods exploit sparse structure but still generate and execute dense trajectories between anchors; the modification is to the \emph{model}, not to the \emph{supervision}. Faster-execution methods~\citep{arachchige2025sail,huang2025daducorki} are closest to SkiP, but adapt execution velocity or the execution pipeline rather than the training target.

\paragraph{Frequency-domain action representations.}
FAST~\citep{pertsch2025fast} uses DCT-based tokenization for autoregressive VLA policies.
FreqPolicy variants explore hierarchical frequency modeling~\citep{zhong2025freqpolicy} and frequency consistency for one-step generation~\citep{su2025freqpolicy}.
Wavelet Policy~\citep{yang2025waveletpolicy} operates in the wavelet domain.
These works use frequency representations as a modeling space for dense trajectory generation.
SkiP uses spectral energy as a \emph{supervisory signal} for temporal abstraction: the frequency content tells us \emph{where} to allocate dense prediction, not how to represent the action itself. We call this procedure Motion Spectrum Keying (\S\ref{sec:method:keyseg}).

\section{Method}
\label{sec:method}

The design separates two questions: \emph{what} information a trajectory region carries and \emph{how} the policy should respond to that region.
Because the two are decoupled, the relabeling mechanism works with any binary temporal annotation, while the labeling procedure can be swapped or improved independently.

\subsection{Problem Setup}
\label{sec:method:setup}
We consider imitation learning from offline robot demonstrations.
Each demonstration is a trajectory $\tau = \{(o_t, a_t)\}_{t=1}^{T}$, where $o_t$ denotes the observation at time $t$ and $a_t \in \mathbb{R}^{d}$ denotes a continuous control command.
Our goal is to learn a policy $\pi_\theta(a \mid o)$ that solves the task while using decision steps efficiently.

Trajectory information density is highly non-uniform: during contacts and precise alignment, the policy must refine actions densely, but during smooth motions, step-by-step prediction is redundant.
SkiP allocates decision steps accordingly, skipping through skip segments and concentrating refinement on key segments.

The framework has two components.
\emph{Action relabeling} (\S\ref{sec:method:relabel}) is the core learning mechanism: given any binary temporal annotation $y_t \in \{0,1\}$ that marks refine-worthy timesteps, it modifies the behavior cloning targets so that a single policy learns both skipping and refinement.
\emph{Motion Spectrum Keying} (MSK, \S\ref{sec:method:keyseg}) is one concrete instantiation of this annotation, based on short-time DCT spectral energy.

\subsection{Action Relabel: Skip-and-Refine Learning}
\label{sec:method:relabel}
Assume we have a set of key segments $\mathcal{S}=\{[s_i,e_i]\}_{i=1}^{M}$ extracted from each demonstration (\S\ref{sec:method:keyseg}; illustrated in Figure~\ref{fig:action_relabel}).
Here $s_i$ and $e_i$ denote the start and end timestep of the $i$-th key segment.
Let $y_t \in \{0,1\}$ be a per-timestep indicator: $y_t=1$ if $t$ lies inside a key segment, $y_t=0$ otherwise.

We train a single policy to realize two behaviors under the same action interface.
When $y_t=1$, the policy should perform dense refinement by predicting the immediate next-step continuation.
When $y_t=0$, the policy should skip toward the next interaction-heavy phase by predicting the action at the \emph{entrance} of the next key segment.

Formally, define the next key-segment start time:
\begin{equation}
t^{+}(t) = \min \{s_i : s_i > t\},
\label{eq:next_start}
\end{equation}
with $t^{+}(t)$ undefined if no future key segment exists.
We construct a relabeled chunk start index:
\begin{equation}
t^\star(t) =
\begin{cases}
t+1, & y_t = 1, \\
t^{+}(t), & y_t = 0 \text{ and } t^{+}(t) \text{ exists}, \\
t+1, & \text{otherwise},
\end{cases}
\label{eq:target_start}
\end{equation}
and use it to form a relabeled target chunk $\tilde{\mathbf{A}}_t \in \mathbb{R}^{H \times d}$ for chunk index $h=1,\ldots,H$, with a padding mask $m_{t,h} \in \{0,1\}$:
\begin{equation}
\begin{aligned}
k_{t,h} &\triangleq t^\star(t) + h - 1, \\
\tilde{a}_{t,h} &=
\begin{cases}
a_{k_{t,h}}, & k_{t,h} \le T, \\
\mathbf{0}, & \text{otherwise},
\end{cases} \\
m_{t,h} &= \mathbb{I}[k_{t,h} \le T],
\end{aligned}
\label{eq:relabel_chunk}
\end{equation}
where $H$ is the chunk length.
We then minimize a masked imitation loss:
\begin{equation}
\mathcal{L}(\theta) = \mathbb{E}_{(o_t,\cdot)\sim\mathcal{D}} \left[ \frac{1}{\sum_{h} m_{t,h}} \sum_{h=1}^{H} m_{t,h} \, \ell(\hat{a}_{t,h},\, \tilde{a}_{t,h}) \right],
\label{eq:bc_loss}
\end{equation}
where $\ell$ is the imitation loss used by the corresponding backbone.
This relabeling unifies skipping and refinement within a single policy.
Inside key segments, $t^\star(t) = t+1$ so the target chunk matches the immediate next-step continuation, training dense feedback-driven refinement.
Outside key segments, $t^\star(t) = t^{+}(t)$ so the target chunk starts at the next key-segment entrance; the policy is trained to jump ahead, implicitly skipping the intermediate skip-segment actions.

\subsection{Motion Spectrum Keying (MSK)}
\label{sec:method:keyseg}
We now describe how the binary annotation $y_t$ is produced.
We introduce \emph{Motion Spectrum Keying} (MSK), which extracts key segments by measuring local frequency content of the action time series.
Prior methods rely on heuristic keyframes such as gripper-state changes or velocity zero-crossings~\citep{james2022c2f-arm}, which capture pick-and-place events but miss sustained high-precision motion like sweeping or dragging.
MSK measures local motion complexity directly.

\paragraph{DCT decomposition.}
Given an action sequence $\{a_t\}_{t=1}^{T}$, we compute per-step velocities $v_t = a_t - a_{t-1}$.
For each $t$, we apply a discrete cosine transform along the temporal axis over a centered length-$W$ window of velocities to obtain spectral coefficients
\begin{equation}
c_{t,k} = \sum_{n=0}^{W-1} \alpha_k \, v_{t,n} \cos\!\left(\frac{\pi(2n+1)k}{2W}\right),
\label{eq:dct}
\end{equation}
where $\alpha_k$ is the standard normalization constant.
We compute spectral energy $E_{t,k} = \|c_{t,k}\|_2^2$ and define the high-frequency energy ratio
\begin{equation}
r_t = \frac{\sum_{k \in \mathcal{H}} E_{t,k}}{\sum_{k} E_{t,k}},
\qquad \mathcal{H} = \{k : k \ge \lceil W/2 \rceil\},
\label{eq:high_ratio}
\end{equation}
where $\mathcal{H}$ is the upper half of the frequency spectrum.
Large $r_t$ indicates rapid local changes associated with contacts or corrections.
We threshold $\{r_t\}$ at the per-episode quantile $q$ and group consecutive positives into segments.

\paragraph{Bend-based augmentation.}
Some task-relevant events appear as brief geometric deviations rather than high-frequency oscillations. We complement the spectral signal with a bend score that compares end-effector translations against a straight-line reference over the same window:
\begin{equation}
b_t = \frac{1}{W} \sum_{n=0}^{W-1} \frac{\|p_{t,n} - \ell_{t,n}\|_2}{\bar{s}_t},
\label{eq:bend}
\end{equation}
where $\ell_{t,n}$ is the linear interpolant between the first and last in-window translations and $\bar{s}_t$ is the mean step length, making $b_t$ scale-invariant. Timesteps with $b_t$ above its per-episode threshold are added to the key segments.

\paragraph{Final partition.}
The key segments are the union of frequency-based segments, bend-based segments, and small neighborhoods around heuristic keyframes (gripper-state changes and end-effector velocity zero-crossings, following~\citet{james2022c2f-arm}); the complement forms the skip segments.
We sweep the window size and quantile threshold in Sec.~\ref{sec:exp:ablation} and ablate the bend and keyframe-union components in Table~\ref{tab:ablate_components}; full implementation details are in Appendix~\ref{sec:appendix:repro}.

\section{Experiments}
\label{sec:exp}

\subsection{Experimental Setup}
\label{sec:exp:setup}

\paragraph{Benchmarks.}
We evaluate SkiP across four settings that vary in policy architecture, observation modality, and embodiment.
On RLBench~\citep{rlbench}, we use 60 manipulation tasks with a 7-DoF Franka Panda, 4 RGB cameras at $128{\times}128$, 100 demonstrations per task, and a transformer-based policy adapted from CoA~\citep{zhang2025coa}.
On RoboMimic~\citep{mandlekar2021robomimic}, we evaluate 4 image-based manipulation tasks using a Diffusion Policy UNet backbone~\citep{chi2023diffusion}.
On RoboTwin~\citep{mu2024robotwin}, we evaluate 8 bimanual tasks with a DP3~\citep{ze2024dp3} backbone in the clean point-cloud setting.
For real-robot evaluation, we fine-tune the pretrained $\pi_{0.5}$~\citep{intelligence2025pi05} on 3 tabletop tasks and include RLBench-18 as a simulation counterpart.

\paragraph{Baselines.}
On RLBench, we compare against DP~\citep{chi2023diffusion}, ACT~\citep{zhao2023act}, Chain-of-Action forward/reverse (CoA-fwd/CoA-rev)~\citep{zhang2025coa}, and a keyframe-only variant (KF-only).
On RoboMimic and RoboTwin, we compare against the respective backbone baselines and CoA variants.

\paragraph{Metrics.}
We report task \emph{success rate} (SR) and execution efficiency measured by average executed control steps per episode (\emph{Steps}).
We additionally report \emph{Steps$_{\text{succ}}$}, the average steps over successful episodes only, to isolate efficiency conditional on completion.

\subsection{Simulation Benchmarks}
\label{sec:exp:simulation}

\begin{figure}[t]
\centering
\includegraphics[width=\columnwidth]{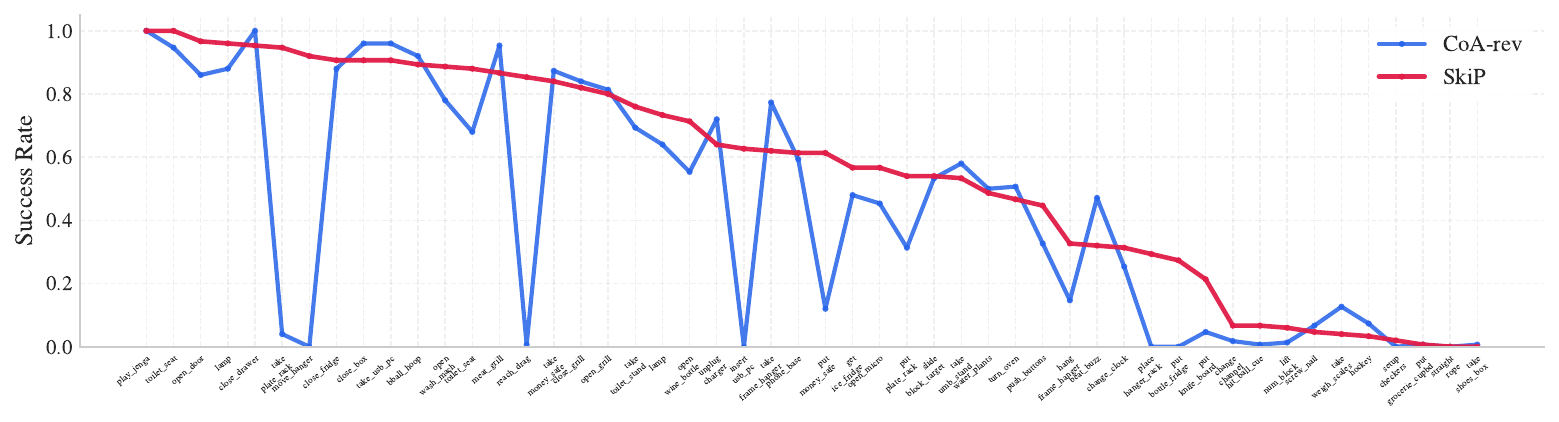}
\caption{Per-task success rates on RLBench-50 (tasks sorted by best SR). SkiP improves over CoA-rev on a broad range of tasks, especially challenging long-horizon manipulations.}
\label{fig:rlbench50}
\end{figure}

\paragraph{RLBench.}
We evaluate on the RLBench-10 task suite used by CoA(Table~\ref{tab:rlbench10_main}) and the broader RLBench-60 suite (Figure~\ref{fig:rlbench50}; aggregate stats in caption).
On RLBench-10, SkiP achieves the highest average SR ($0.850$) while requiring the fewest executed Steps ($72.9$) among all methods, improving over CoA-rev by $+0.149$ SR and $54.6$ fewer steps.
KF-only attains very low Steps$_{\text{succ}}$ ($10.9$) but unreliable SR ($0.493$).

\newcommand{\bC}[2]{\cellcolor{skipColorSteps!#1!white}#2}

\begin{table}[t]
\centering
\scriptsize
\setlength{\tabcolsep}{1pt}
\renewcommand{\arraystretch}{1.10}
\caption{RLBench-10 per-task SR$\uparrow$ and Steps$\downarrow$ (mean$\pm$std over three runs). Blue saturation: darker = better. $\dagger$\,SkiP variant that replaces MSK with key segments from Gemini-2.5-Pro~\citep{gemini2025}. Best in \textbf{bold}, second \underline{underlined}.}
\label{tab:rlbench10_main}
\begin{tabular}{l *{12}{C{1.0cm}}}
\toprule
\multirow{2}{*}{Method}
 & \multicolumn{2}{c}{Avg} & \multicolumn{2}{c}{Overall}
 & \multicolumn{2}{c}{\Task{open-box}} & \multicolumn{2}{c}{\Task{open-drawer}}
 & \multicolumn{2}{c}{\Task{pick-up-cup}} & \multicolumn{2}{c}{\Task{press-switch}} \\
 \cmidrule(lr){2-3} \cmidrule(lr){4-5} \cmidrule(lr){6-7} \cmidrule(lr){8-9} \cmidrule(lr){10-11} \cmidrule(lr){12-13}
 & SR$\uparrow$ & Steps$\downarrow$ & Steps$_{\text{s}}\!\downarrow$ & rk$\downarrow$
 & SR & Steps & SR & Steps & SR & Steps & SR & Steps \\
\midrule
DP & \bC{5}{\ms{0.43}{.02}} & \bC{10}{\ms{160.0}{3}} & \bC{10}{\ms{119.1}{2}} & \bC{5}{5.4} & \bC{10}{\ms{0.34}{.02}} & \bC{5}{\ms{201.5}{4}} & \bC{5}{\ms{0.26}{.06}} & \bC{5}{\ms{147.0}{6}} & \bC{5}{\ms{0.35}{.10}} & \bC{5}{\ms{214.7}{25}} & \bC{15}{\ms{0.45}{.03}} & \bC{32}{\underline{\ms{146.9}{10}}} \\
KF-only & \bC{10}{\ms{0.49}{.01}} & \bC{32}{\underline{\ms{113.1}{3}}} & \bC{45}{\textbf{\ms{10.9}{3}}} & \bC{10}{4.3} & \bC{5}{\ms{0.17}{.03}} & \bC{10}{\ms{198.0}{7}} & \bC{10}{\ms{0.55}{.07}} & \bC{32}{\underline{\ms{82.1}{13}}} & \bC{32}{\underline{\ms{0.69}{.05}}} & \bC{45}{\textbf{\ms{94.4}{15}}} & \bC{10}{\ms{0.34}{.06}} & \bC{15}{\ms{187.0}{9}} \\
CoA-fwd & \bC{15}{\ms{0.68}{.01}} & \bC{5}{\ms{161.5}{1}} & \bC{5}{\ms{133.3}{1}} & \bC{15}{3.5} & \bC{22}{\ms{0.64}{.00}} & \bC{15}{\ms{185.8}{0}} & \bC{22}{\ms{0.81}{.02}} & \bC{10}{\ms{119.4}{1}} & \bC{15}{\ms{0.56}{.03}} & \bC{10}{\ms{194.3}{9}} & \bC{5}{\ms{0.28}{.00}} & \bC{5}{\ms{243.8}{0}} \\
ACT & \bC{32}{\underline{\ms{0.71}{.01}}} & \bC{15}{\ms{139.4}{1}} & \bC{15}{\ms{106.1}{1}} & \bC{32}{\underline{3.1}} & \bC{15}{\ms{0.53}{.03}} & \bC{22}{\ms{181.2}{2}} & \bC{45}{\textbf{\ms{1.00}{.00}}} & \bC{22}{\ms{90.8}{0}} & \bC{10}{\ms{0.55}{.04}} & \bC{15}{\ms{174.2}{10}} & \bC{45}{\textbf{\ms{0.58}{.03}}} & \bC{22}{\ms{177.1}{4}} \\
CoA-rev & \bC{22}{\ms{0.70}{.05}} & \bC{22}{\ms{127.5}{5}} & \bC{22}{\ms{86.2}{3}} & \bC{22}{3.1} & \bC{32}{\underline{\ms{0.72}{.35}}} & \bC{32}{\underline{\ms{158.4}{31}}} & \bC{15}{\ms{0.79}{.03}} & \bC{15}{\ms{101.5}{6}} & \bC{22}{\ms{0.68}{.03}} & \bC{22}{\ms{138.3}{7}} & \bC{22}{\ms{0.46}{.02}} & \bC{10}{\ms{191.3}{5}} \\
SkiP$^\dagger$ & \bC{5}{\ms{0.30}{.01}} & \bC{5}{\ms{183.1}{2}} & \bC{5}{\ms{97.6}{7}} & \bC{5}{6.3} & \bC{5}{\ms{0.24}{.07}} & \bC{5}{\ms{208.5}{9}} & \bC{5}{\ms{0.23}{.02}} & \bC{5}{\ms{165.7}{1}} & \bC{5}{\ms{0.13}{.05}} & \bC{5}{\ms{272.1}{9}} & \bC{5}{\ms{0.35}{.05}} & \bC{5}{\ms{200.2}{8}} \\
SkiP & \bC{45}{\textbf{\ms{0.85}{.01}}} & \bC{45}{\textbf{\ms{72.9}{2}}} & \bC{32}{\underline{\ms{43.4}{1}}} & \bC{45}{\textbf{1.6}} & \bC{45}{\textbf{\ms{0.91}{.02}}} & \bC{45}{\textbf{\ms{87.1}{7}}} & \bC{32}{\underline{\ms{1.00}{.00}}} & \bC{45}{\textbf{\ms{39.8}{2}}} & \bC{45}{\textbf{\ms{0.74}{.06}}} & \bC{32}{\underline{\ms{96.2}{17}}} & \bC{32}{\underline{\ms{0.54}{.03}}} & \bC{45}{\textbf{\ms{144.4}{7}}} \\
\bottomrule
\end{tabular}

\vspace{0.6ex}
\begin{tabular}{l *{12}{C{1.0cm}}}
\toprule
\multirow{2}{*}{Method}
 & \multicolumn{2}{c}{\Task{push-button}} & \multicolumn{2}{c}{\Task{reach-target}}
 & \multicolumn{2}{c}{\Task{stack-wine}} & \multicolumn{2}{c}{\Task{sweep-dustpan}}
 & \multicolumn{2}{c}{\Task{take-lid-off}} & \multicolumn{2}{c}{\Task{turn-tap}} \\
 \cmidrule(lr){2-3} \cmidrule(lr){4-5} \cmidrule(lr){6-7} \cmidrule(lr){8-9} \cmidrule(lr){10-11} \cmidrule(lr){12-13}
 & SR & Steps & SR & Steps & SR & Steps & SR & Steps & SR & Steps & SR & Steps \\
\midrule
DP & \bC{15}{\ms{0.66}{.04}} & \bC{10}{\ms{130.8}{4}} & \bC{5}{\ms{0.60}{.03}} & \bC{15}{\ms{57.6}{4}} & \bC{5}{\ms{0.40}{.04}} & \bC{5}{\ms{216.9}{5}} & \bC{10}{\ms{0.08}{.03}} & \bC{5}{\ms{165.4}{2}} & \bC{5}{\ms{0.73}{.06}} & \bC{10}{\ms{127.8}{8}} & \bC{5}{\ms{0.44}{.03}} & \bC{5}{\ms{191.2}{10}} \\
KF-only & \bC{10}{\ms{0.64}{.03}} & \bC{32}{\underline{\ms{95.2}{9}}} & \bC{45}{\textbf{\ms{0.72}{.00}}} & \bC{45}{\textbf{\ms{37.2}{0}}} & \bC{10}{\ms{0.48}{.06}} & \bC{32}{\underline{\ms{97.4}{10}}} & \bC{5}{\ms{0.04}{.00}} & \bC{10}{\ms{163.4}{0}} & \bC{22}{\ms{0.80}{.00}} & \bC{32}{\underline{\ms{48.5}{1}}} & \bC{10}{\ms{0.50}{.03}} & \bC{32}{\underline{\ms{127.3}{8}}} \\
CoA-fwd & \bC{32}{\underline{\ms{0.97}{.04}}} & \bC{15}{\ms{126.6}{3}} & \bC{15}{\ms{0.68}{.00}} & \bC{10}{\ms{68.7}{0}} & \bC{15}{\ms{0.52}{.00}} & \bC{10}{\ms{213.5}{1}} & \bC{32}{\underline{\ms{0.96}{.00}}} & \bC{15}{\ms{120.2}{0}} & \bC{10}{\ms{0.73}{.02}} & \bC{5}{\ms{159.5}{2}} & \bC{32}{\underline{\ms{0.64}{.03}}} & \bC{10}{\ms{183.6}{8}} \\
ACT & \bC{5}{\ms{0.58}{.07}} & \bC{5}{\ms{162.5}{5}} & \bC{32}{\underline{\ms{0.72}{.00}}} & \bC{22}{\ms{46.6}{4}} & \bC{32}{\underline{\ms{0.90}{.02}}} & \bC{15}{\ms{163.6}{2}} & \bC{22}{\ms{0.88}{.00}} & \bC{22}{\ms{111.8}{0}} & \bC{15}{\ms{0.80}{.03}} & \bC{15}{\ms{113.1}{4}} & \bC{22}{\ms{0.57}{.05}} & \bC{15}{\ms{172.8}{7}} \\
CoA-rev & \bC{22}{\ms{0.88}{.05}} & \bC{22}{\ms{100.8}{11}} & \bC{22}{\ms{0.69}{.02}} & \bC{5}{\ms{70.5}{3}} & \bC{22}{\ms{0.69}{.03}} & \bC{22}{\ms{146.6}{5}} & \bC{15}{\ms{0.73}{.40}} & \bC{32}{\underline{\ms{105.7}{35}}} & \bC{32}{\underline{\ms{0.83}{.02}}} & \bC{22}{\ms{103.2}{4}} & \bC{15}{\ms{0.54}{.03}} & \bC{22}{\ms{158.4}{6}} \\
SkiP$^\dagger$ & \bC{5}{\ms{0.65}{.04}} & \bC{10}{\ms{118.8}{8}} & \bC{5}{\ms{0.56}{.00}} & \bC{5}{\ms{74.5}{0}} & \bC{5}{\ms{0.03}{.02}} & \bC{5}{\ms{244.6}{4}} & \bC{5}{\ms{0.15}{.02}} & \bC{5}{\ms{159.8}{2}} & \bC{5}{\ms{0.33}{.02}} & \bC{5}{\ms{188.3}{6}} & \bC{5}{\ms{0.37}{.02}} & \bC{5}{\ms{198.3}{6}} \\
SkiP & \bC{45}{\textbf{\ms{0.98}{.02}}} & \bC{45}{\textbf{\ms{26.2}{4}}} & \bC{10}{\ms{0.68}{.00}} & \bC{32}{\underline{\ms{45.3}{0}}} & \bC{45}{\textbf{\ms{1.00}{.00}}} & \bC{45}{\textbf{\ms{77.3}{0}}} & \bC{45}{\textbf{\ms{1.00}{.00}}} & \bC{45}{\textbf{\ms{59.4}{0}}} & \bC{45}{\textbf{\ms{0.97}{.02}}} & \bC{45}{\textbf{\ms{33.4}{3}}} & \bC{45}{\textbf{\ms{0.68}{.03}}} & \bC{45}{\textbf{\ms{120.0}{8}}} \\
\bottomrule
\end{tabular}
\end{table}

\begin{figure}[t]
\centering
\begin{subfigure}[t]{0.48\columnwidth}
\centering
\includegraphics[width=\linewidth]{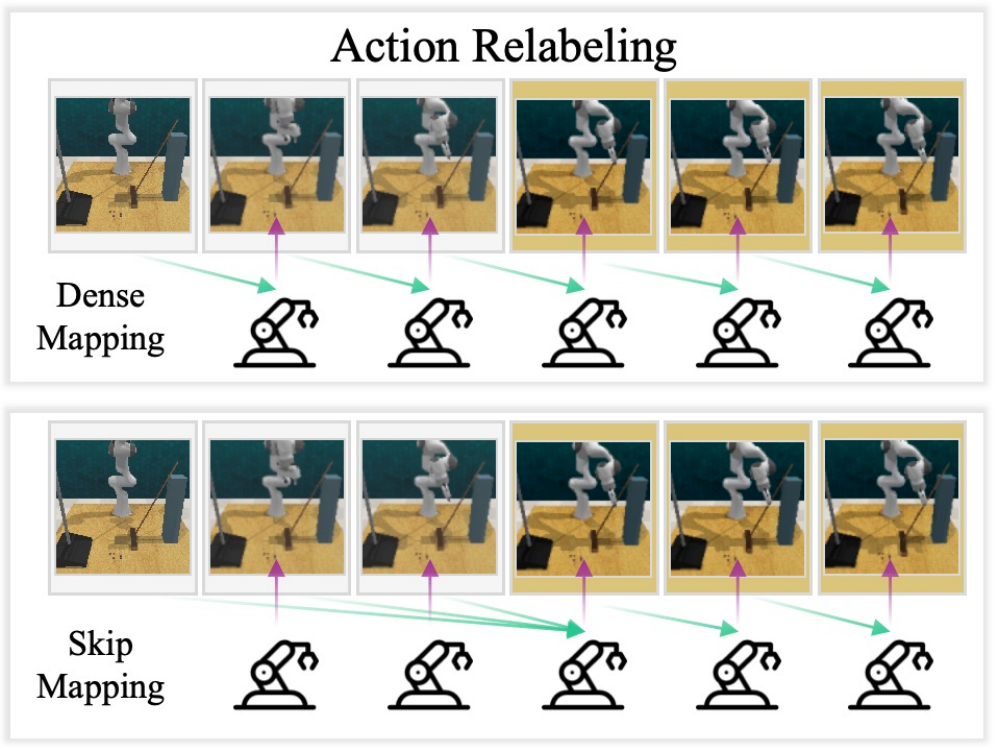}
\caption{Action relabeling. Top: standard dense mapping (every timestep predicts the next). Bottom: skip mapping (skip-segment timesteps predict the next key-segment entrance).}
\label{fig:action_relabel}
\end{subfigure}
\hfill
\begin{subfigure}[t]{0.48\columnwidth}
\centering
\includegraphics[width=\linewidth]{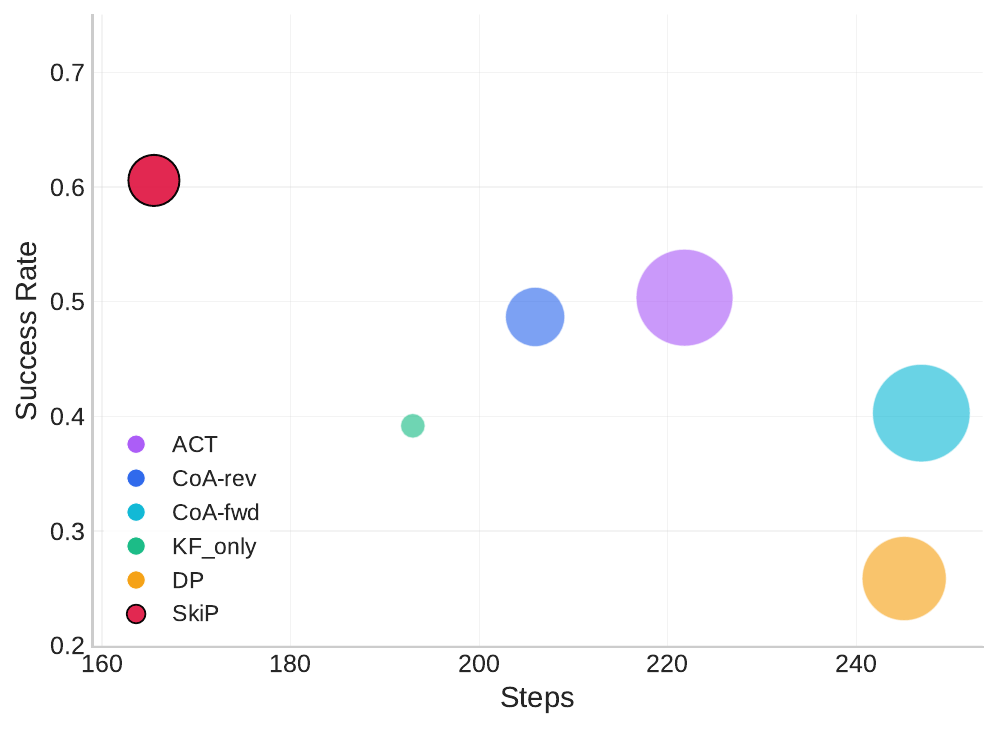}
\caption{SR vs.\ Steps on RLBench-60. Bubble area encodes Steps$_{\text{succ}}$ (smaller = more efficient).}
\label{fig:scatter}
\end{subfigure}
\caption{(a) Illustration of SkiP's relabeling scheme: in skip segments, the training target jumps to the next key-segment entrance. (b) SkiP sits in the upper-left: highest SR, fewest steps.}
\label{fig:relabel_scatter}
\end{figure}

Figure~\ref{fig:relabel_scatter}(b) plots the SR--Steps trade-off across all 60 RLBench tasks: SkiP sits in the upper-left corner with both the highest SR and the fewest steps, and the smallest Steps$_{\text{succ}}$ bubble among methods at comparable SR.

Figure~\ref{fig:rlbench50} compares SkiP with CoA-rev on the 50 tasks outside RLBench-10, sorted by difficulty.
SkiP improves success on a broad range of tasks, especially long-horizon tasks where concentrating control steps around key segments matters most.

\paragraph{RoboMimic.}
Table~\ref{tab:robomimic} reports results on RoboMimic using a Diffusion Policy UNet backbone in the image observation setting.
SkiP achieves the highest average SR, surpasses CoA-rev on the hardest task \Task{square}, and reduces Steps$_\text{succ}$ by $32\%$ relative to CoA-rev.

\begin{table}[t]
\centering
\small
\setlength{\tabcolsep}{3pt}
\renewcommand{\arraystretch}{1.10}
\caption{RoboMimic results with Diffusion Policy UNet backbone in the image observation setting.}
\label{tab:robomimic}
\begin{tabular}{l *{4}{C{1.7cm}}}
\toprule
Task & CoA-rev & CoA-fwd & SkiP$^\dagger$ & SkiP \\
\midrule
\Task{lift}       & \bC{15}{\ms{0.960}{0.016}}              & \bC{45}{\textbf{\ms{1.000}{0.000}}} & \bC{5}{\ms{0.013}{0.019}} & \bC{45}{\textbf{\ms{1.000}{0.000}}} \\
\Task{can}        & \bC{45}{\textbf{\ms{0.880}{0.016}}}     & \bC{30}{\underline{\ms{0.873}{0.034}}} & \bC{5}{\ms{0.233}{0.009}} & \bC{15}{\ms{0.827}{0.019}} \\
\Task{square}     & \bC{30}{\underline{\ms{0.420}{0.016}}}  & \bC{15}{\ms{0.327}{0.025}}              & \bC{5}{\ms{0.247}{0.019}} & \bC{45}{\textbf{\ms{0.673}{0.062}}} \\
\Task{transport}  & \bC{45}{\textbf{\ms{0.633}{0.057}}}     & \bC{15}{\ms{0.427}{0.047}}              & \bC{5}{\ms{0.000}{0.000}} & \bC{30}{\underline{\ms{0.587}{0.041}}} \\
\midrule
\textbf{Avg SR}$\uparrow$              & \bC{30}{\underline{\ms{0.723}{0.013}}} & \bC{15}{\ms{0.657}{0.013}} & \bC{5}{\ms{0.123}{0.002}} & \bC{45}{\textbf{\ms{0.772}{0.012}}} \\
\textbf{Steps$_\text{succ}$}$\downarrow$ & \bC{15}{\ms{211.8}{3.2}}    & \bC{5}{\ms{211.9}{4.2}}   & \bC{45}{\textbf{\ms{88.6}{3.9}}} & \bC{30}{\underline{\ms{144.1}{2.2}}} \\
\bottomrule
\end{tabular}
\end{table}

\paragraph{RoboTwin.}
Table~\ref{tab:robotwin} reports results on 8 bimanual tasks using a DP3 backbone with clean point cloud demonstrations.
SkiP achieves the highest average SR, with the VLM variant SkiP$^\dagger$ close behind, and substantially reduces Steps$_{\text{succ}}$ relative to CoA-rev and DP3.

\begin{table}[t]
\centering
\scriptsize
\setlength{\tabcolsep}{2pt}
\renewcommand{\arraystretch}{1.10}
\caption{RoboTwin 2.0 results with DP3 backbone in the clean point-cloud setting.}
\label{tab:robotwin}
\begin{tabular}{l *{5}{C{1.8cm}}}
\toprule
Task & DP3 & CoA-fwd & CoA-rev & SkiP$^\dagger$ & SkiP \\
\midrule
\TaskUS{adjust\_bottle}          & \bC{25}{\ms{0.987}{0.009}}              & \bC{15}{\ms{0.980}{0.008}}              & \bC{15}{\ms{0.980}{0.008}}              & \bC{45}{\textbf{\ms{0.997}{0.005}}} & \bC{35}{\underline{\ms{0.993}{0.005}}} \\
\TaskUS{beat\_block\_hammer}     & \bC{5}{\ms{0.627}{0.050}}               & \bC{15}{\ms{0.770}{0.029}}              & \bC{25}{\ms{0.817}{0.025}}              & \bC{45}{\textbf{\ms{0.857}{0.021}}} & \bC{35}{\underline{\ms{0.833}{0.024}}} \\
\TaskUS{handover\_block}         & \bC{5}{\ms{0.713}{0.025}}               & \bC{35}{\underline{\ms{0.840}{0.024}}} & \bC{15}{\ms{0.780}{0.029}}              & \bC{25}{\ms{0.790}{0.014}}              & \bC{45}{\textbf{\ms{0.870}{0.008}}} \\
\TaskUS{move\_can\_pot}          & \bC{35}{\underline{\ms{0.527}{0.012}}} & \bC{15}{\ms{0.440}{0.037}}              & \bC{5}{\ms{0.393}{0.060}}               & \bC{25}{\ms{0.487}{0.078}}              & \bC{45}{\textbf{\ms{0.540}{0.014}}} \\
\TaskUS{open\_microwave}         & \bC{5}{\ms{0.297}{0.076}}               & \bC{15}{\ms{0.593}{0.012}}              & \bC{25}{\ms{0.683}{0.031}}              & \bC{45}{\textbf{\ms{0.860}{0.036}}} & \bC{35}{\underline{\ms{0.830}{0.113}}} \\
\TaskUS{place\_container\_plate} & \bC{5}{\ms{0.767}{0.009}}               & \bC{15}{\ms{0.857}{0.012}}              & \bC{45}{\textbf{\ms{0.883}{0.019}}}    & \bC{35}{\underline{\ms{0.863}{0.005}}} & \bC{35}{\underline{\ms{0.863}{0.012}}} \\
\TaskUS{place\_empty\_cup}       & \bC{5}{\ms{0.637}{0.019}}               & \bC{35}{\underline{\ms{0.893}{0.012}}} & \bC{45}{\textbf{\ms{0.910}{0.028}}}    & \bC{25}{\ms{0.833}{0.045}}              & \bC{15}{\ms{0.810}{0.029}} \\
\TaskUS{place\_shoe}             & \bC{5}{\ms{0.360}{0.024}}               & \bC{15}{\ms{0.437}{0.069}}              & \bC{45}{\textbf{\ms{0.443}{0.082}}}    & \bC{25}{\ms{0.440}{0.008}}              & \bC{45}{\textbf{\ms{0.443}{0.075}}} \\
\midrule
\textbf{Avg SR}$\uparrow$                & \bC{5}{\ms{0.614}{0.005}}  & \bC{15}{\ms{0.726}{0.013}} & \bC{25}{\ms{0.736}{0.019}}              & \bC{35}{\underline{\ms{0.766}{0.017}}} & \bC{45}{\textbf{\ms{0.773}{0.016}}} \\
\textbf{Steps$_\text{succ}$}$\downarrow$ & \bC{5}{\ms{268.7}{4.8}}    & \bC{15}{\ms{242.7}{6.1}}   & \bC{25}{\ms{176.2}{3.5}}                & \bC{45}{\textbf{\ms{121.7}{2.3}}}   & \bC{35}{\underline{\ms{126.6}{7.3}}} \\
\bottomrule
\end{tabular}
\end{table}

\paragraph{Summary.}
Across three simulation benchmarks, the same training-target modification transfers across DP, DP3, and autoregressive backbones without per-backbone tuning.
We also compare MSK with VLM-derived key segments from Gemini-2.5-Pro (SkiP$^\dagger$).
VLM segments are competitive on RoboTwin, but perform much worse on RLBench and RoboMimic, suggesting that semantic phase boundaries are not always precise enough for contact-rich relabeling.

\subsection{VLA Fine-tuning}
\label{sec:exp:vla}

\begin{wraptable}{r}{0.46\textwidth}
\centering
\scriptsize
\setlength{\tabcolsep}{3pt}
\renewcommand{\arraystretch}{1.0}
\vspace{-1.5em}
\caption{Real-robot $\pi_{0.5}$ fine-tuning results. Time = wall-clock time (min:sec).}
\label{tab:real_robot}
\begin{tabular}{l cccc}
\toprule
 & Base & KF-only & CoA & SkiP \\
\midrule
\multicolumn{5}{l}{\textit{SR (\%)$\uparrow$}} \\
\Task{pour-water}   & \bC{32}{\underline{40.0}} & \bC{10}{6.7}  & \bC{22}{33.3} & \bC{45}{\textbf{46.7}} \\
\Task{stack-bowls}  & \bC{32}{\underline{33.3}} & \bC{10}{6.7}  & \bC{22}{26.7} & \bC{45}{\textbf{53.3}} \\
\Task{tidy-up-desk} & \bC{32}{\underline{66.7}} & \bC{10}{13.3} & \bC{22}{40.0} & \bC{45}{\textbf{73.3}} \\
\midrule
\multicolumn{5}{l}{\textit{Steps$\downarrow$}} \\
\Task{pour-water}   & \bC{22}{290.4} & \bC{10}{309.6} & \bC{32}{\underline{281.9}} & \bC{45}{\textbf{265.4}} \\
\Task{stack-bowls}  & \bC{22}{246.3} & \bC{10}{271.4} & \bC{32}{\underline{226.8}} & \bC{45}{\textbf{204.5}} \\
\Task{tidy-up-desk} & \bC{22}{250.7} & \bC{10}{286.8} & \bC{32}{\underline{232.4}} & \bC{45}{\textbf{207.4}} \\
\midrule
\multicolumn{5}{l}{\textit{Time$\downarrow$}} \\
\Task{pour-water}   & \bC{22}{3:44} & \bC{10}{4:01} & \bC{32}{\underline{3:39}} & \bC{45}{\textbf{3:28}} \\
\Task{stack-bowls}  & \bC{22}{2:41} & \bC{10}{2:59} & \bC{32}{\underline{2:29}} & \bC{45}{\textbf{2:16}} \\
\Task{tidy-up-desk} & \bC{22}{2:20} & \bC{10}{2:43} & \bC{32}{\underline{2:12}} & \bC{45}{\textbf{2:00}} \\
\bottomrule
\end{tabular}
\vspace{-1em}
\end{wraptable}

We fine-tune $\pi_{0.5}$~\citep{intelligence2025pi05}, a successor to the $\pi_0$ vision-language-action flow model~\citep{black2024pi0} trained on the Open X-Embodiment dataset~\citep{openx2023rtx}, using both standard behavior cloning (\textbf{Base}) and SkiP's relabeling objective to check whether the approach works with pretrained foundation models.
Since $\pi_{0.5}$ operates in joint space, we discover key segments from end-effector pose trajectories and transfer the resulting partition to joint-space action targets.

\paragraph{Real-robot results.}
Table~\ref{tab:real_robot} reports results on three tabletop tasks with 15 rollouts each, comparing SkiP against three baselines: standard behavior cloning (Base), keyframe-only prediction (KF-only), and Chain-of-Action (CoA).
SkiP gets the highest SR on all three tasks and uses fewer executed steps and less wall-clock time per episode.
On \Task{stack-bowls}, SR rises from $33.3\%$ (Base) to $53.3\%$ (SkiP) while wall-clock time drops from $2$\,m\,$41$\,s to $2$\,m\,$16$\,s.
KF-only is fast on successful runs but rarely succeeds in the open-world setting, with SR consistently below $14\%$.
That SkiP improves a pretrained VLA suggests relabeling is complementary to large-scale pretraining: the pretrained weights provide general motor competence, and relabeling teaches the policy where to spend its decision budget.

\paragraph{Simulation counterpart: RLBench-18.}
We repeat the same protocol on RLBench-18 in simulation.
Averaged across 18 tasks (3 seeds each), SkiP improves average SR from $18.59\%$ to $20.96\%$ ($+2.37$ points) while reducing Steps$_\text{succ}$ from $108.30$ to $66.38$ ($-39\%$).
The gains come from tasks where $\pi_{0.5}$ already achieves nonzero success under standard fine-tuning; tasks at $0\%$ SR remain unsolved by both methods, so SkiP helps most when the policy has partial competence to begin with.
Full per-task results are in Appendix~\ref{sec:appendix:rlbench18}.

\subsection{Ablations}
\label{sec:exp:ablation}

\begin{wrapfigure}{r}{0.46\textwidth}
\centering
\vspace{-1.2em}
\includegraphics[width=0.44\textwidth]{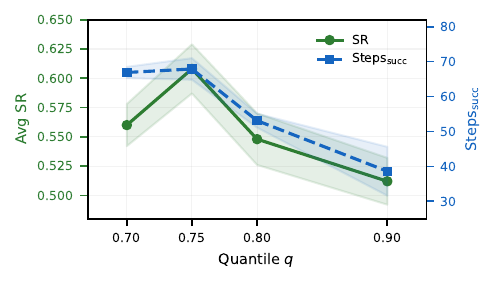}
\vspace{-1ex}
\caption{Ablation on quantile threshold $q$ (RLBench-60, 3 eval repeats, shaded bands $= \pm 1$ std).
SR peaks at $q{=}0.75$ and drops for $q \geq 0.80$; Steps$_{\text{succ}}$ decreases monotonically.}
\label{fig:ablate_hq}
\vspace{-1em}
\end{wrapfigure}

\paragraph{Quantile threshold $q$.}
The threshold $q$ controls how conservatively key segments are labeled: larger $q$ marks fewer timesteps as refine-worthy.
Figure~\ref{fig:ablate_hq} traces SR and Steps$_\text{succ}$ as $q$ varies from $0.70$ to $0.90$ on RLBench-60.
SR peaks at $q{=}0.75$ and degrades for $q \geq 0.80$, where the policy skips too aggressively and misses critical manipulation segments.
Steps$_\text{succ}$ decreases monotonically with $q$ as fewer timesteps are marked refine-worthy.
We select $q{=}0.75$ as the default, which achieves the best SR while already providing substantial step reduction.

\paragraph{ST-DCT window size $W$.}
We further vary the short-time DCT window $W$ while keeping other hyperparameters fixed (Table~\ref{tab:ablate_window}).
$W{=}16$ gives the best SR; smaller windows ($W{=}4,8$) lose temporal context and reduce SR by $5$--$8$ points, while $W{=}32$ produces the lowest Steps$_{\text{succ}}$ but at the cost of SR (overly selective key-segment signal that omits some corrective motions).

\paragraph{Key-segment components.}
Table~\ref{tab:ablate_components} ablates keyframe union ($u$) and bend post-processing ($b$).
Removing both reduces SR from $0.606$ to $0.530$; the remaining $0.530$ with DCT-only labels is still above CoA-rev's $0.490$ on RLBench-60.

\begin{table}[t]
\centering
\small
\begin{minipage}[t]{0.42\textwidth}
\centering
\setlength{\tabcolsep}{5pt}
\captionof{table}{Ablation on the ST-DCT window size $W$ (RLBench-60).}
\label{tab:ablate_window}
\begin{tabular}{lccc}
\toprule
$W$ & SR$\uparrow$ & Steps$\downarrow$ & Steps$_{\text{succ}}\downarrow$ \\
\midrule
4  & 0.529 & 178.4 & \underline{50.6} \\
8  & 0.547 & 175.6 & 57.2 \\
16 & \textbf{0.606} & \textbf{165.5} & 64.1 \\
32 & 0.514 & \underline{176.7} & \textbf{47.2} \\
\bottomrule
\end{tabular}
\end{minipage}\hfill
\begin{minipage}[t]{0.55\textwidth}
\centering
\setlength{\tabcolsep}{5pt}
\captionof{table}{Ablation of keyframe union ($u$) and bend post-processing ($b$) on RLBench-60.}
\label{tab:ablate_components}
\begin{tabular}{lcccc}
\toprule
Variant & $u$ & $b$ & SR$\uparrow$ & Steps$\downarrow$ \\
\midrule
w/o $u$ \& $b$ &  &  & 0.530 & 176.3 \\
w/o union &  & \cmark & \underline{0.574} & 168.8 \\
w/o bend  & \cmark &  & \underline{0.574} & \textbf{163.0} \\
full      & \cmark & \cmark & \textbf{0.606} & \underline{165.5} \\
\bottomrule
\end{tabular}
\end{minipage}
\end{table}

\paragraph{Label source comparison.}
To test whether the benefit of action relabeling depends on \emph{how} key segments are identified, we train three alternative segmentation strategies on RLBench-10 with the same architecture and hyperparameters, changing only the label source:
\emph{Random Stride} (RS) places key segments at fixed periodic intervals matching the ${\sim}25\%$ key ratio of MSK;
\emph{Velocity Only} (VO) labels high-velocity timesteps as key ($q{=}0.75$ quantile threshold on raw velocity magnitude);
\emph{Low Velocity Key} (LV) inverts this, labeling low-velocity timesteps as key under the intuition that slow motion corresponds to careful manipulation.
Table~\ref{tab:label_source} shows that MSK outperforms all alternatives ($+0.078$ over LV, $+0.331$ over RS).
RS performs poorly despite matching the key-segment ratio, showing that segment \emph{placement} matters more than count.
VO suffers catastrophic failures on contact-heavy tasks (\Task{open-box}: $0.027$, \Task{sweep-dustpan}: $0.040$).
LV is the strongest alternative but still misses trajectory curvature patterns that MSK captures.

\begin{table}[t]
\centering
\small
\setlength{\tabcolsep}{5pt}
\caption{Label source ablation on RLBench-10 (SR$\uparrow$). Same architecture; only segmentation differs.}
\label{tab:label_source}
\begin{tabular}{l cccc}
\toprule
Task & RS & VO & LV & MSK (SkiP) \\
\midrule
\Task{open-box}       & \bC{25}{\ms{.76}{.00}} & \bC{5}{\ms{.03}{.02}}  & \bC{25}{\ms{.76}{.00}} & \bC{45}{\textbf{\ms{.91}{.02}}} \\
\Task{open-drawer}    & \bC{15}{\ms{.72}{.00}} & \bC{5}{\ms{.04}{.00}}  & \bC{35}{\underline{\ms{.96}{.00}}} & \bC{45}{\textbf{\ms{1.0}{.00}}} \\
\Task{pick-up-cup}    & \bC{5}{\ms{.23}{.05}}  & \bC{35}{\underline{\ms{.75}{.04}}} & \bC{45}{\textbf{\ms{.81}{.08}}} & \bC{25}{\ms{.76}{.06}} \\
\Task{press-switch}   & \bC{5}{\ms{.41}{.05}}  & \bC{45}{\textbf{\ms{.79}{.04}}} & \bC{25}{\ms{.57}{.05}} & \bC{15}{\ms{.53}{.04}} \\
\Task{push-button}    & \bC{5}{\ms{.12}{.00}}  & \bC{15}{\ms{.65}{.02}} & \bC{45}{\textbf{\ms{1.0}{.00}}} & \bC{35}{\underline{\ms{.99}{.02}}} \\
\Task{reach-target}   & \bC{25}{\ms{.64}{.00}} & \bC{45}{\textbf{\ms{.68}{.00}}} & \bC{5}{\ms{.31}{.02}}  & \bC{45}{\textbf{\ms{.68}{.00}}} \\
\Task{stack-wine}     & \bC{5}{\ms{.77}{.02}}  & \bC{15}{\ms{.85}{.02}} & \bC{25}{\ms{.88}{.00}} & \bC{45}{\textbf{\ms{1.0}{.00}}} \\
\Task{sweep-dustpan}  & \bC{15}{\ms{.64}{.00}} & \bC{5}{\ms{.04}{.00}}  & \bC{45}{\textbf{\ms{1.0}{.00}}} & \bC{45}{\textbf{\ms{1.0}{.00}}} \\
\Task{take-lid-off}   & \bC{5}{\ms{.60}{.00}}  & \bC{45}{\textbf{\ms{.99}{.02}}} & \bC{15}{\ms{.71}{.02}} & \bC{35}{\underline{\ms{.97}{.02}}} \\
\Task{turn-tap}       & \bC{5}{\ms{.31}{.04}}  & \bC{15}{\ms{.64}{.03}} & \bC{45}{\textbf{\ms{.73}{.02}}} & \bC{35}{\ms{.67}{.02}} \\
\midrule
\textbf{Avg}          & \bC{5}{.520} & \bC{15}{.545} & \bC{25}{.773} & \bC{45}{\textbf{.851}} \\
\bottomrule
\end{tabular}
\end{table}

\subsection{Analysis: Does SkiP Learn to Skip and Refine?}
\label{sec:exp:analysis}

\begin{figure}[t]
\centering
\includegraphics[width=\columnwidth]{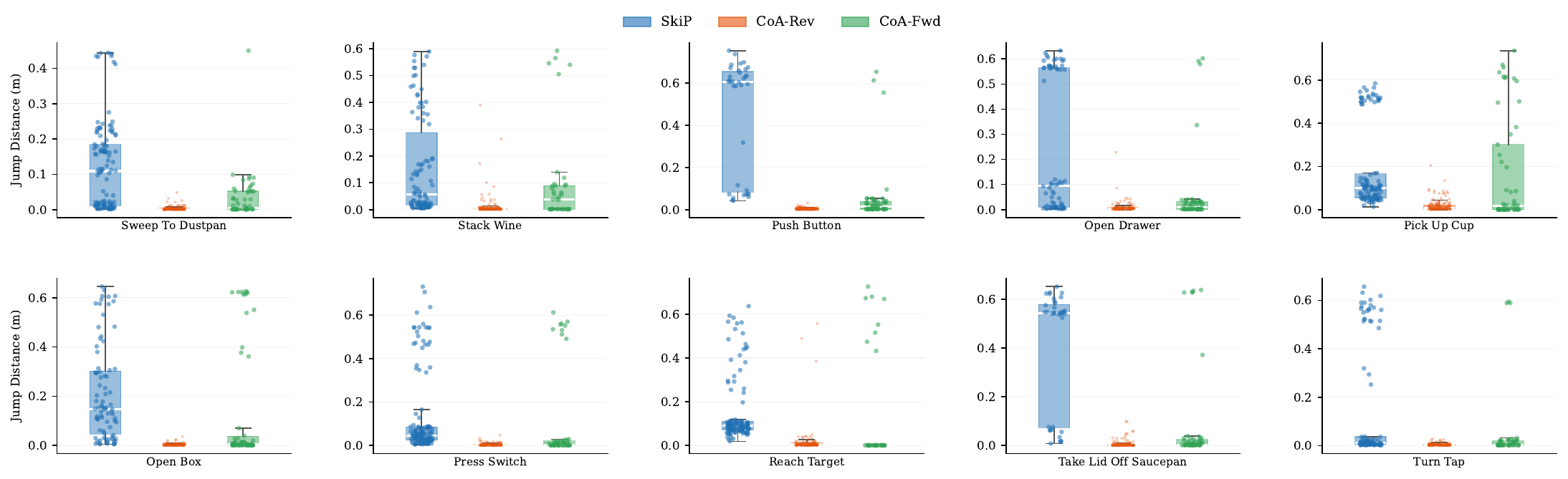}
\vspace{-3ex}
\caption{Action displacement distribution per policy call across 10 RLBench tasks. SkiP shows a bimodal pattern: large jumps in skip mode, small adjustments in key mode. CoA-rev concentrates near zero; CoA-fwd shows moderate unimodal displacements.}
\label{fig:action_disp}
\end{figure}

We measure the \emph{jump distance} $\|a_{1} - p_{\text{ee}}\|_2$ per policy call during evaluation, splitting SkiP calls into \emph{key} and \emph{skip} categories by a per-task displacement threshold.
Figure~\ref{fig:action_disp} shows a clear bimodal pattern across $10$ RLBench tasks: skip-mode calls produce jumps of $0.1$--$0.7$\,m, while key-mode calls cluster near zero.
CoA-rev concentrates near zero; CoA-fwd shows moderate unimodal displacements.
This supports the interpretation that relabeling teaches distinct output modes rather than a single averaged behavior.

\section{Conclusion}
\label{sec:conclusion}

We introduced SkiP, which learns when to skip and when to refine by relabeling behavior cloning targets.
Across RLBench, RoboMimic, RoboTwin, and real-robot $\pi_{0.5}$ fine-tuning, SkiP cuts executed steps by $15$--$40\%$ while matching or improving success rates across various policy architectures.
The broader message: temporal resolution in imitation learning can be controlled at the supervision level; changing \emph{what} the policy predicts is sufficient to produce adaptive skip-and-refine behavior.
We discuss limitations (absolute-target requirement, failure modes, label source dependence) and detailed failure analysis in Appendix~\ref{sec:appendix:limitations} and~\ref{sec:appendix:failure}.

\clearpage

\bibliography{references}
\bibliographystyle{plainnat}

\newpage
\appendix

\section{SkiP Relabeling Pseudocode}
\label{sec:appendix:pseudocode}
Algorithm~\ref{alg:skip_relabel} summarizes the relabeling logic for constructing SkiP training targets as $H$-step chunks.
\begin{algorithm}[htbp]
\caption{SkiP training sample construction}
\label{alg:skip_relabel}
\begin{algorithmic}[1]
\REQUIRE Demonstration actions $\{a_t\}_{t=1}^{T}$, \texttt{region}[1..T], \texttt{next\_high}[1..T] (set to $\bot$ if no future key segment), chunk length $H$
\ENSURE Index $i$, target chunk $\tilde{\mathbf{A}}\in\mathbb{R}^{H\times d}$, mask $m\in\{0,1\}^{H}$
\IF{\texttt{region}[$i$]$=0$ \AND \texttt{next\_high}[$i$]$\neq\bot$}
\STATE $t_{\text{tgt}}\leftarrow$ \texttt{next\_high}[$i$] \hfill\COMMENT{\bf skip sample}
\ELSE
\STATE $t_{\text{tgt}}\leftarrow i+1$ \hfill\COMMENT{\bf refine, or skip-tail with no future key segment}
\ENDIF
\FOR{$h=1$ {\bf to} $H$}
\IF{$t_{\text{tgt}}+h-1 \le T$}
\STATE $\tilde{\mathbf{A}}[h] \leftarrow a_{t_{\text{tgt}}+h-1}$; \;\; $m[h]\leftarrow 1$
\ELSE
\STATE $\tilde{\mathbf{A}}[h] \leftarrow \mathbf{0}$; \;\; $m[h]\leftarrow 0$
\ENDIF
\ENDFOR
\end{algorithmic}
\end{algorithm}

\section{Reproducibility Details}
\label{sec:appendix:repro}

\subsection{Evaluation Protocol}
Each policy query outputs an $H$-step action chunk, and the environment executes the full chunk before the next query.
We disable temporal ensembling for all methods.
Steps counts the number of executed \texttt{env.step} calls until termination.
All settings use absolute target actions: RLBench and RoboMimic use absolute end-effector pose targets, while RoboTwin and the $\pi_{0.5}$ setup use absolute joint position targets.
For RoboMimic, \Task{tool\_hang} is excluded from the main average because all methods obtain $0.00$ SR on this task.

\subsection{Policy Architecture}
SkiP uses the same transformer-based policy architecture as CoA on RLBench (ResNet-18 image encoder, 4-layer encoder / 6-layer decoder transformer, $d_\text{model}{=}512$, 78.4M parameters total).
On RoboMimic, we use Diffusion Policy's UNet architecture.
On RoboTwin, we use DP3's 3D diffusion architecture.
On real-robot and RLBench-18, we fine-tune from the $\pi_{0.5}$ checkpoint.

\subsection{Motion Spectrum Keying Details}
\label{sec:appendix:msk_details}
MSK has two tunable hyperparameters: the DCT window size $W{=}16$ (centered, even split $[t{-}8,t{+}7]$) and the quantile threshold $q{=}0.75$ (per-episode percentile on the high-frequency energy ratio); both are swept in \S\ref{sec:exp:ablation}.
The upper half of the DCT spectrum ($k \ge \lceil W/2 \rceil$) defines the high-frequency band, and segments shorter than $3$ steps are discarded as noise.
The remaining implementation constants are fixed across all experiments without tuning:
bend-based augmentation uses a deviation cutoff of $0.30$ with $\pm 2$-step expansion;
heuristic keyframes (gripper-state changes and velocity zero-crossings, following~\citet{james2022c2f-arm}) are included with a $5$-step neighborhood;
the first $20\%$ of each episode is excluded to avoid labeling the initial reset motion.

\section{Efficiency Decomposition}
\label{sec:appendix:efficiency}

Table~\ref{tab:eff_decomp} decomposes executed steps and policy forward calls on RLBench-10.
SkiP achieves the lowest executed steps (72.9) with only 3.57 policy calls per episode.
CoA-fwd has the fewest calls (2.08) due to very long chunks, but many more executed steps (160.5).
CoA-rev's adaptive early-stop mechanism often collapses effective chunk length to 1 step after the first interval, inflating calls to 41.36 per episode.

\begin{table}[t]
\centering
\small
\setlength{\tabcolsep}{7pt}
\caption{Efficiency decomposition on RLBench-10.}
\label{tab:eff_decomp}
\begin{tabular}{lcc}
\toprule
Method & Steps$\downarrow$ & \# forward calls/ep$\downarrow$ \\
\midrule
DP & 160.0 & 8.34 \\
ACT & 138.8 & 7.49 \\
KF-only & \underline{113.4} & 111.94 \\
CoA-fwd & 160.5 & \textbf{2.08} \\
CoA-rev & 125.4 & 41.36 \\
SkiP & \textbf{72.5} & \underline{3.57} \\
\bottomrule
\end{tabular}
\end{table}

\section{RLBench-18 Fine-tuning from $\pi_{0.5}$}
\label{sec:appendix:rlbench18}

Table~\ref{tab:rlbench18_pi05} reports per-task success rate and Steps$_{\text{succ}}$ when fine-tuning $\pi_{0.5}$ on the RLBench-18 suite.
SkiP improves SR from $18.59\%$ to $20.96\%$ on average while reducing Steps$_{\text{succ}}$ from $108.30$ to $66.38$ ($-39\%$).
Gains concentrate on tasks where the base policy achieves nonzero success (e.g., \TaskUS{close\_jar}, \TaskUS{meat\_off\_grill}, \TaskUS{reach\_and\_drag}); tasks at $0\%$ SR for both methods likely require stronger initialization or additional data.

\begin{table}[htbp]
\centering
\scriptsize
\setlength{\tabcolsep}{3pt}
\renewcommand{\arraystretch}{1.08}
\caption{RLBench-18 fine-tuning from $\pi_{0.5}$ (3 seeds). Bold indicates better performance.}
\label{tab:rlbench18_pi05}
\begin{tabular}{p{3.8cm} l C{1.2cm} C{1.8cm}}
\toprule
Task & Method & SR$\uparrow$ & Steps$_{\text{succ}}\downarrow$ \\
\midrule
\multirow{2}{*}{\TaskUS{close\_jar}} & Base & 18.67$\pm$6.11 & 189.77$\pm$21.21 \\
 & SkiP & \textbf{42.67$\pm$2.31} & \textbf{140.49$\pm$11.32} \\
\midrule
\multirow{2}{*}{\TaskUS{insert\_onto\_square\_peg}} & Base & 0.00$\pm$0.00 & --- \\
 & SkiP & 0.00$\pm$0.00 & --- \\
\midrule
\multirow{2}{*}{\TaskUS{light\_bulb\_in}} & Base & 0.00$\pm$0.00 & --- \\
 & SkiP & 0.00$\pm$0.00 & --- \\
\midrule
\multirow{2}{*}{\TaskUS{meat\_off\_grill}} & Base & 36.00$\pm$6.93 & 150.61$\pm$15.41 \\
 & SkiP & \textbf{72.00$\pm$4.00} & \textbf{94.79$\pm$2.97} \\
\midrule
\multirow{2}{*}{\TaskUS{open\_drawer}} & Base & \textbf{48.00$\pm$6.93} & 101.32$\pm$3.53 \\
 & SkiP & 32.00$\pm$8.00 & \textbf{47.26$\pm$10.67} \\
\midrule
\multirow{2}{*}{\TaskUS{place\_cups}} & Base & 0.00$\pm$0.00 & --- \\
 & SkiP & 0.00$\pm$0.00 & --- \\
\midrule
\multirow{2}{*}{\TaskUS{place\_shape\_in\_shape\_sorter}} & Base & 0.00$\pm$0.00 & --- \\
 & SkiP & 0.00$\pm$0.00 & --- \\
\midrule
\multirow{2}{*}{\TaskUS{place\_wine\_at\_rack\_location}} & Base & \textbf{48.00$\pm$13.86} & 209.21$\pm$1.78 \\
 & SkiP & 25.33$\pm$9.24 & \textbf{99.94$\pm$9.82} \\
\midrule
\multirow{2}{*}{\TaskUS{push\_buttons}} & Base & \textbf{24.00$\pm$4.00} & 76.36$\pm$1.94 \\
 & SkiP & 9.33$\pm$8.33 & \textbf{29.50$\pm$25.75} \\
\midrule
\multirow{2}{*}{\TaskUS{put\_groceries\_in\_cupboard}} & Base & 0.00$\pm$0.00 & --- \\
 & SkiP & 0.00$\pm$0.00 & --- \\
\midrule
\multirow{2}{*}{\TaskUS{put\_item\_in\_drawer}} & Base & \textbf{25.33$\pm$4.62} & 457.98$\pm$8.88 \\
 & SkiP & 5.33$\pm$6.11 & \textbf{256.33$\pm$245.79} \\
\midrule
\multirow{2}{*}{\TaskUS{put\_money\_in\_safe}} & Base & 20.00$\pm$10.58 & 238.81$\pm$3.11 \\
 & SkiP & \textbf{36.00$\pm$6.93} & \textbf{195.97$\pm$18.91} \\
\midrule
\multirow{2}{*}{\TaskUS{reach\_and\_drag}} & Base & 21.33$\pm$8.33 & 182.56$\pm$26.19 \\
 & SkiP & \textbf{49.33$\pm$6.11} & \textbf{127.49$\pm$5.22} \\
\midrule
\multirow{2}{*}{\TaskUS{slide\_block\_to\_color\_target}} & Base & 18.67$\pm$4.62 & 84.64$\pm$0.63 \\
 & SkiP & \textbf{37.33$\pm$2.31} & \textbf{47.26$\pm$12.71} \\
\midrule
\multirow{2}{*}{\TaskUS{stack\_blocks}} & Base & 0.00$\pm$0.00 & --- \\
 & SkiP & 0.00$\pm$0.00 & --- \\
\midrule
\multirow{2}{*}{\TaskUS{stack\_cups}} & Base & 0.00$\pm$0.00 & --- \\
 & SkiP & 0.00$\pm$0.00 & --- \\
\midrule
\multirow{2}{*}{\TaskUS{sweep\_to\_dustpan\_of\_size}} & Base & \textbf{48.00$\pm$4.00} & 122.66$\pm$7.08 \\
 & SkiP & 45.33$\pm$2.31 & \textbf{61.84$\pm$4.95} \\
\midrule
\multirow{2}{*}{\TaskUS{turn\_tap}} & Base & \textbf{26.67$\pm$8.33} & 135.56$\pm$5.37 \\
 & SkiP & 22.67$\pm$2.31 & \textbf{94.06$\pm$11.08} \\
\midrule
\multirow{2}{*}{Overall (avg)} & Base & 18.59$\pm$1.05 & 108.30$\pm$1.63 \\
 & SkiP & \textbf{20.96$\pm$1.51} & \textbf{66.38$\pm$11.61} \\
\bottomrule
\end{tabular}
\end{table}

\section{Real-Robot Rollout Examples}
\label{sec:appendix:real_robot_examples}

\begin{figure}[htbp]
\centering
\includegraphics[width=\textwidth]{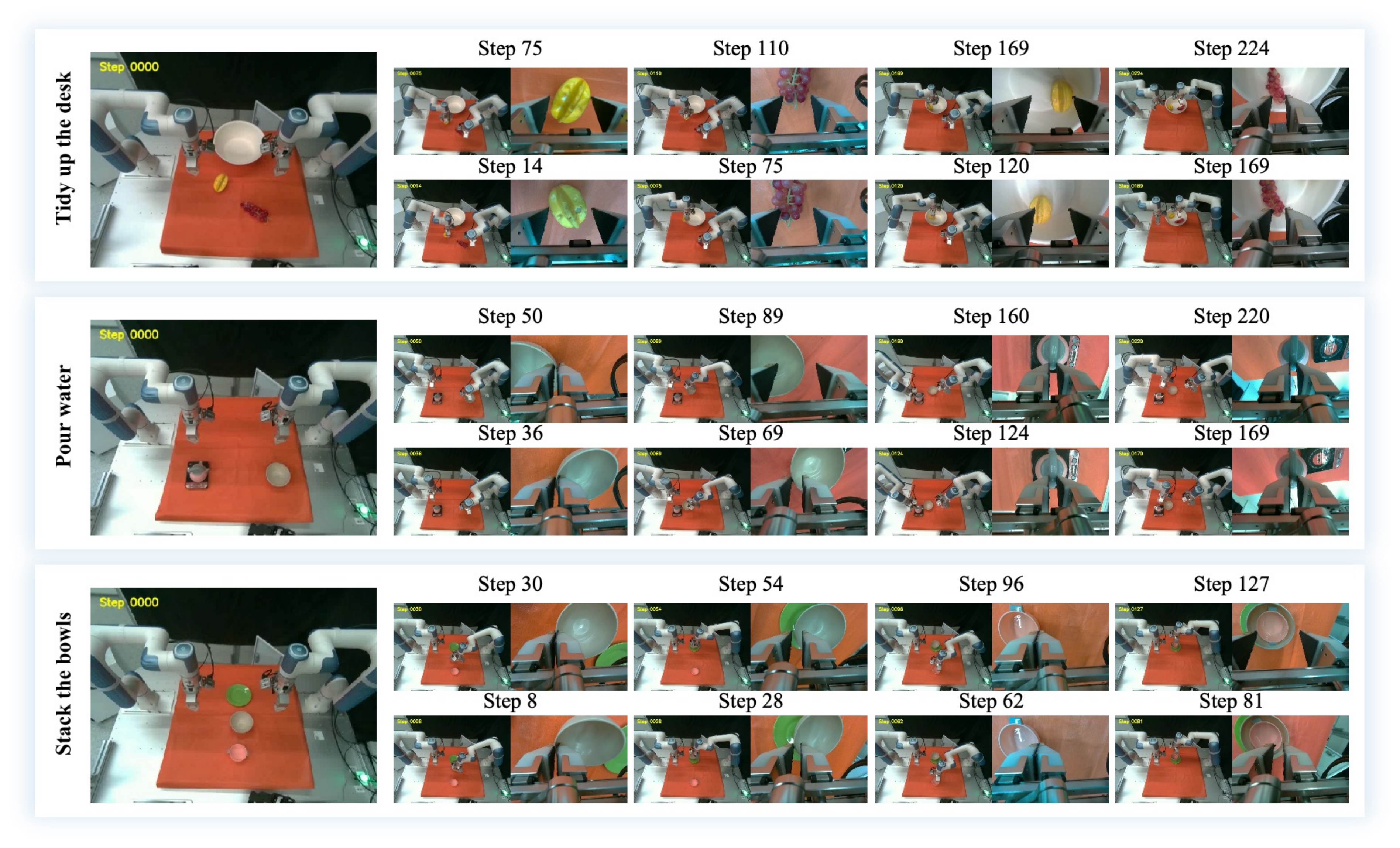}
\caption{Real-robot rollout examples for \Task{pour-water}, \Task{stack-bowls}, and \Task{tidy-up-desk}.}
\label{fig:real_robot_examples}
\end{figure}

Figure~\ref{fig:real_robot_examples} shows the real-robot setup and rollout examples for the three tabletop tasks.
These examples match the tasks used in Table~\ref{tab:real_robot}.

\section{RLBench Per-Task Results}
\label{sec:appendix:rlbench}

We report extended per-task results on the RLBench-60 suite. The main paper's Table~\ref{tab:rlbench10_main} already covers RLBench-10 SR and Steps; here we provide Steps$_{\text{succ}}$ per task and the full RLBench-50 numbers.

\subsection{RLBench-10 Per-Task Steps$_{\text{succ}}$}
\label{sec:appendix:rlbench10_steps_succ}

Table~\ref{tab:rlbench10_steps_succ} reports per-task Steps$_{\text{succ}}$ (averaged over successful episodes only).
KF-only is extremely efficient on pick-and-place-like tasks where it succeeds (e.g., \TaskUS{reach\_target}, \TaskUS{push\_button}, \TaskUS{take\_lid\_off}) but fails outright on contact-heavy tasks like \TaskUS{open\_box} and \TaskUS{sweep\_to\_dustpan}, motivating reporting Steps alongside Steps$_{\text{succ}}$.

\begin{table}[htbp]
\centering
\scriptsize
\setlength{\tabcolsep}{3pt}
\renewcommand{\arraystretch}{1.05}
\caption{RLBench-10 per-task Steps$_{\text{succ}}$ (average executed steps over successful episodes). Dashes indicate $0$ successful episodes for that method on that task.}
\label{tab:rlbench10_steps_succ}
\begin{tabular}{l cccccc}
\toprule
Task & DP & KF-only & CoA-fwd & ACT & CoA-rev & SkiP \\
\midrule
\TaskUS{open\_box}        & 165.2 & \underline{71.1} & 163.7 & 143.1 & 132.7 & \textbf{73.8} \\
\TaskUS{open\_drawer}     & 96.1  & \textbf{2.0}  & 105.6 & 90.8  & 87.9  & \underline{39.7} \\
\TaskUS{pick\_up\_cup}    & 97.6  & \textbf{2.1}  & 117.8 & 78.4  & 76.1  & \underline{24.6} \\
\TaskUS{press\_switch}    & 134.6 & \textbf{14.0} & 176.6 & 121.7 & 110.1 & \underline{40.3} \\
\TaskUS{push\_button}     & 91.3  & \textbf{8.4}  & 124.6 & 99.3  & 80.4  & \underline{21.7} \\
\TaskUS{reach\_target}    & 43.7  & \textbf{1.1}  & 39.9  & 38.5  & 45.4  & \underline{5.4} \\
\TaskUS{stack\_wine}      & 186.3 & \textbf{2.1}  & 201.4 & 160.9 & 111.4 & \underline{77.3} \\
\TaskUS{sweep\_to\_dustpan} & 118.5 & \textbf{4.0} & 118.1 & 103.9 & 82.1  & \underline{59.5} \\
\TaskUS{take\_lid\_off}   & 108.2 & \textbf{3.5}  & 133.4 & 89.8  & 85.1  & \underline{27.2} \\
\TaskUS{turn\_tap}        & 150.4 & \textbf{2.4}  & 145.3 & 129.5 & 87.1  & \underline{61.6} \\
\midrule
\textbf{Avg}              & 119.2 & \textbf{11.1} & 132.6 & 105.6 & 89.8 & \underline{43.1} \\
\bottomrule
\end{tabular}
\end{table}

\subsection{Per-Task Chunk Length $H$}
\label{sec:appendix:chunk_length}

Table~\ref{tab:chunk_length} reports the per-task action-chunk length $H$ used by CoA-fwd/rev (max sub-trajectory length between consecutive keyframes) versus SkiP (max discovered key-segment length). Across RLBench-60, CoA's median $H$ is $138$ versus SkiP's $31$, a $4.5\times$ gap. This gap explains why CoA-fwd attains low policy-call counts under open-loop execution while SkiP retains shorter chunks to prioritize fine-grained control near key segments.

\begin{table}[htbp]
\centering
\scriptsize
\setlength{\tabcolsep}{4pt}
\caption{Per-task chunk length $H$ on RLBench-60. CoA's $H$ is the per-task training action-sequence length; SkiP's $H$ is the longest discovered key-segment length under default MSK settings ($W{=}16$, $q{=}0.75$).}
\label{tab:chunk_length}
\begin{tabular}{l cc | l cc}
\toprule
Task & $H$ (CoA) & $H$ (SkiP) & Task & $H$ (CoA) & $H$ (SkiP) \\
\midrule
\TaskUS{basketball\_in\_hoop} & 190 & 25 & \TaskUS{open\_microwave}        & 126 & 43 \\
\TaskUS{beat\_the\_buzz}      & 150 & 27 & \TaskUS{open\_washing\_machine} & 149 & 33 \\
\TaskUS{change\_channel}      & 134 & 31 & \TaskUS{open\_wine\_bottle}     & 113 & 32 \\
\TaskUS{change\_clock}        & 211 & 35 & \TaskUS{phone\_on\_base}        & 149 & 41 \\
\TaskUS{close\_box}           & 211 & 23 & \TaskUS{pick\_up\_cup}          & 118 & 30 \\
\TaskUS{close\_drawer}        & 97  & 15 & \TaskUS{place\_hanger\_on\_rack} & 167 & 34 \\
\TaskUS{close\_fridge}        & 185 & 38 & \TaskUS{play\_jenga}            & 86  & 12 \\
\TaskUS{close\_grill}         & 108 & 23 & \TaskUS{press\_switch}          & 145 & 28 \\
\TaskUS{get\_ice\_from\_fridge} & 242 & 44 & \TaskUS{take\_shoes\_out\_of\_box} & 192 & 18 \\
\TaskUS{hang\_frame\_on\_hanger} & 245 & 28 & \TaskUS{toilet\_seat\_down}      & 117 & 27 \\
\TaskUS{hit\_ball\_with\_queue} & 210 & 33 & \TaskUS{turn\_tap}              & 119 & 28 \\
\TaskUS{hockey}               & 134 & 26 & \TaskUS{water\_plants}          & 156 & 35 \\
\TaskUS{insert\_usb\_in\_computer} & 97 & 16 & \multicolumn{3}{c}{\textit{(remaining tasks omitted for space)}} \\
\TaskUS{lamp\_off}            & 101 & 31 & & & \\
\TaskUS{lamp\_on}             & 102 & 31 & \textbf{Median (60 tasks)} & \textbf{138} & \textbf{31} \\
\bottomrule
\end{tabular}
\end{table}

\subsection{RLBench-50 Per-Task Results}

Table~\ref{tab:rlbench50_full} reports per-task SR and Steps for all 50 remaining tasks.
SkiP frequently reduces Steps on tasks with long free-space motion while maintaining competitive SR.
Notable examples: on \TaskUS{close\_fridge}, SkiP reduces Steps from 105.2 (CoA-rev) to 51.1 while improving SR; on \TaskUS{lamp\_off}, SkiP reaches 0.960 SR with 26.7 Steps vs.\ CoA-rev's 0.880 SR with 78.8 Steps.

\addtocounter{table}{1}

\begingroup
\scriptsize
\setlength{\tabcolsep}{1.4pt}
\renewcommand{\arraystretch}{1.05}
\begin{longtable}{@{}c@{}}
\caption{RLBench-50 per-task SR$\uparrow$ and Steps$\downarrow$. Blue saturation: darker = better. Best in \textbf{bold}, second \underline{underlined}.}
\label{tab:rlbench50_full}\\
\begin{tabular}{l*{5}{C{1.05cm}C{1.15cm}}}
\toprule
\multirow{2}{*}{Method} & \multicolumn{2}{C{2.20cm}}{\TaskUS{basketball\_in\_hoop}} & \multicolumn{2}{C{2.20cm}}{\TaskUS{beat\_the\_buzz}} & \multicolumn{2}{C{2.20cm}}{\TaskUS{change\_channel}} & \multicolumn{2}{C{2.20cm}}{\TaskUS{change\_clock}} & \multicolumn{2}{C{2.20cm}}{\TaskUS{close\_box}} \\
 & SR$\uparrow$ & Steps$\downarrow$ & SR$\uparrow$ & Steps$\downarrow$ & SR$\uparrow$ & Steps$\downarrow$ & SR$\uparrow$ & Steps$\downarrow$ & SR$\uparrow$ & Steps$\downarrow$ \\
\midrule
ACT & \bC{5}{0.827} & \bC{5}{159.1} & \bC{5}{0.253} & \bC{5}{161.7} & \bC{5}{0.027} & \bC{5}{343.8} & \bC{5}{0.247} & \bC{5}{256.9} & \bC{45}{\textbf{0.987}} & \bC{5}{184.4} \\
CoA-r & \bC{45}{\textbf{0.920}} & \bC{45}{\textbf{89.4}} & \bC{45}{\textbf{0.467}} & \bC{25}{\underline{145.2}} & \bC{5}{0.027} & \bC{45}{\textbf{329.6}} & \bC{25}{\underline{0.267}} & \bC{25}{\underline{233.3}} & \bC{25}{\underline{0.960}} & \bC{25}{\underline{120.4}} \\
SkiP & \bC{25}{\underline{0.867}} & \bC{25}{\underline{112.4}} & \bC{25}{\underline{0.320}} & \bC{45}{\textbf{105.0}} & \bC{45}{\textbf{0.093}} & \bC{25}{\underline{330.9}} & \bC{45}{\textbf{0.307}} & \bC{45}{\textbf{219.3}} & \bC{5}{0.933} & \bC{45}{\textbf{92.2}} \\
\bottomrule
\end{tabular}\tabularnewline[0.6ex]
\begin{tabular}{l*{5}{C{1.05cm}C{1.15cm}}}
\toprule
\multirow{2}{*}{Method} & \multicolumn{2}{C{2.20cm}}{\TaskUS{close\_drawer}} & \multicolumn{2}{C{2.20cm}}{\TaskUS{close\_fridge}} & \multicolumn{2}{C{2.20cm}}{\TaskUS{close\_grill}} & \multicolumn{2}{C{2.20cm}}{\TaskUS{get\_ice\_from\_fridge}} & \multicolumn{2}{C{2.20cm}}{\TaskUS{hang\_frame\_on\_hanger}} \\
 & SR$\uparrow$ & Steps$\downarrow$ & SR$\uparrow$ & Steps$\downarrow$ & SR$\uparrow$ & Steps$\downarrow$ & SR$\uparrow$ & Steps$\downarrow$ & SR$\uparrow$ & Steps$\downarrow$ \\
\midrule
ACT & \bC{5}{0.920} & \bC{5}{80.5} & \bC{5}{0.873} & \bC{5}{121.0} & \bC{5}{0.640} & \bC{5}{122.6} & \bC{5}{0.033} & \bC{5}{378.3} & \bC{25}{\underline{0.300}} & \bC{5}{329.9} \\
CoA-r & \bC{45}{\textbf{1.000}} & \bC{25}{\underline{66.6}} & \bC{25}{\underline{0.880}} & \bC{25}{\underline{105.2}} & \bC{45}{\textbf{0.840}} & \bC{25}{\underline{69.3}} & \bC{25}{\underline{0.480}} & \bC{25}{\underline{269.9}} & \bC{5}{0.173} & \bC{25}{\underline{329.8}} \\
SkiP & \bC{25}{\underline{0.947}} & \bC{45}{\textbf{25.2}} & \bC{45}{\textbf{0.893}} & \bC{45}{\textbf{51.1}} & \bC{25}{\underline{0.800}} & \bC{45}{\textbf{67.9}} & \bC{45}{\textbf{0.613}} & \bC{45}{\textbf{263.1}} & \bC{45}{\textbf{0.373}} & \bC{45}{\textbf{281.6}} \\
\bottomrule
\end{tabular}\tabularnewline[0.6ex]
\begin{tabular}{l*{5}{C{1.05cm}C{1.15cm}}}
\toprule
\multirow{2}{*}{Method} & \multicolumn{2}{C{2.20cm}}{\TaskUS{hit\_ball\_with\_queue}} & \multicolumn{2}{C{2.20cm}}{\TaskUS{hockey}} & \multicolumn{2}{C{2.20cm}}{\TaskUS{insert\_usb\_in\_computer}} & \multicolumn{2}{C{2.20cm}}{\TaskUS{lamp\_off}} & \multicolumn{2}{C{2.20cm}}{\TaskUS{lamp\_on}} \\
 & SR$\uparrow$ & Steps$\downarrow$ & SR$\uparrow$ & Steps$\downarrow$ & SR$\uparrow$ & Steps$\downarrow$ & SR$\uparrow$ & Steps$\downarrow$ & SR$\uparrow$ & Steps$\downarrow$ \\
\midrule
ACT & \bC{5}{0.000} & \bC{5}{280.0} & \bC{5}{0.027} & \bC{5}{356.4} & \bC{25}{\underline{0.600}} & \bC{25}{\underline{167.7}} & \bC{5}{0.773} & \bC{5}{106.8} & \bC{25}{\underline{0.667}} & \bC{5}{120.0} \\
CoA-r & \bC{25}{\underline{0.013}} & \bC{25}{\underline{276.9}} & \bC{45}{\textbf{0.067}} & \bC{45}{\textbf{352.0}} & \bC{5}{0.000} & \bC{5}{210.0} & \bC{25}{\underline{0.880}} & \bC{25}{\underline{78.8}} & \bC{5}{0.640} & \bC{25}{\underline{108.9}} \\
SkiP & \bC{45}{\textbf{0.093}} & \bC{45}{\textbf{250.1}} & \bC{5}{0.027} & \bC{25}{\underline{354.3}} & \bC{45}{\textbf{0.613}} & \bC{45}{\textbf{105.9}} & \bC{45}{\textbf{0.960}} & \bC{45}{\textbf{26.7}} & \bC{45}{\textbf{0.747}} & \bC{45}{\textbf{68.9}} \\
\bottomrule
\end{tabular}\tabularnewline[0.6ex]
\begin{tabular}{l*{5}{C{1.05cm}C{1.15cm}}}
\toprule
\multirow{2}{*}{Method} & \multicolumn{2}{C{2.20cm}}{\TaskUS{lift\_numbered\_block}} & \multicolumn{2}{C{2.20cm}}{\TaskUS{meat\_off\_grill}} & \multicolumn{2}{C{2.20cm}}{\TaskUS{move\_hanger}} & \multicolumn{2}{C{2.20cm}}{\TaskUS{open\_door}} & \multicolumn{2}{C{2.20cm}}{\TaskUS{open\_grill}} \\
 & SR$\uparrow$ & Steps$\downarrow$ & SR$\uparrow$ & Steps$\downarrow$ & SR$\uparrow$ & Steps$\downarrow$ & SR$\uparrow$ & Steps$\downarrow$ & SR$\uparrow$ & Steps$\downarrow$ \\
\midrule
ACT & \bC{5}{0.007} & \bC{5}{288.7} & \bC{5}{0.807} & \bC{5}{156.3} & \bC{25}{\underline{0.800}} & \bC{25}{\underline{179.8}} & \bC{25}{\underline{0.953}} & \bC{5}{158.8} & \bC{5}{0.667} & \bC{5}{200.2} \\
CoA-r & \bC{25}{\underline{0.027}} & \bC{25}{\underline{284.3}} & \bC{45}{\textbf{0.947}} & \bC{45}{\textbf{77.4}} & \bC{5}{0.000} & \bC{5}{220.0} & \bC{5}{0.880} & \bC{25}{\underline{142.2}} & \bC{45}{\textbf{0.827}} & \bC{25}{\underline{141.1}} \\
SkiP & \bC{45}{\textbf{0.040}} & \bC{45}{\textbf{279.7}} & \bC{25}{\underline{0.853}} & \bC{25}{\underline{102.5}} & \bC{45}{\textbf{0.920}} & \bC{45}{\textbf{95.0}} & \bC{45}{\textbf{0.973}} & \bC{45}{\textbf{88.5}} & \bC{25}{\underline{0.800}} & \bC{45}{\textbf{123.9}} \\
\bottomrule
\end{tabular}\tabularnewline[0.6ex]
\begin{tabular}{l*{5}{C{1.05cm}C{1.15cm}}}
\toprule
\multirow{2}{*}{Method} & \multicolumn{2}{C{2.20cm}}{\TaskUS{open\_microwave}} & \multicolumn{2}{C{2.20cm}}{\TaskUS{open\_washing\_machine}} & \multicolumn{2}{C{2.20cm}}{\TaskUS{open\_wine\_bottle}} & \multicolumn{2}{C{2.20cm}}{\TaskUS{phone\_on\_base}} & \multicolumn{2}{C{2.20cm}}{\TaskUS{place\_hanger\_on\_rack}} \\
 & SR$\uparrow$ & Steps$\downarrow$ & SR$\uparrow$ & Steps$\downarrow$ & SR$\uparrow$ & Steps$\downarrow$ & SR$\uparrow$ & Steps$\downarrow$ & SR$\uparrow$ & Steps$\downarrow$ \\
\midrule
ACT & \bC{5}{0.393} & \bC{5}{192.3} & \bC{5}{0.580} & \bC{5}{208.2} & \bC{25}{\underline{0.567}} & \bC{5}{146.1} & \bC{5}{0.293} & \bC{5}{288.3} & \bC{25}{\underline{0.060}} & \bC{25}{\underline{320.3}} \\
CoA-r & \bC{25}{\underline{0.467}} & \bC{25}{\underline{170.4}} & \bC{25}{\underline{0.760}} & \bC{25}{\underline{142.8}} & \bC{5}{0.547} & \bC{25}{\underline{135.4}} & \bC{25}{\underline{0.547}} & \bC{25}{\underline{206.1}} & \bC{5}{0.000} & \bC{5}{330.0} \\
SkiP & \bC{45}{\textbf{0.573}} & \bC{45}{\textbf{144.2}} & \bC{45}{\textbf{0.893}} & \bC{45}{\textbf{69.8}} & \bC{45}{\textbf{0.787}} & \bC{45}{\textbf{116.7}} & \bC{45}{\textbf{0.587}} & \bC{45}{\textbf{199.3}} & \bC{45}{\textbf{0.307}} & \bC{45}{\textbf{261.9}} \\
\bottomrule
\end{tabular}\tabularnewline[0.6ex]
\begin{tabular}{l*{5}{C{1.05cm}C{1.15cm}}}
\toprule
\multirow{2}{*}{Method} & \multicolumn{2}{C{2.20cm}}{\TaskUS{play\_jenga}} & \multicolumn{2}{C{2.20cm}}{\TaskUS{push\_buttons}} & \multicolumn{2}{C{2.20cm}}{\TaskUS{put\_bottle\_in\_fridge}} & \multicolumn{2}{C{2.20cm}}{\TaskUS{put\_groceries\_in\_cupboard}} & \multicolumn{2}{C{2.20cm}}{\TaskUS{put\_knife\_on\_chop\_board}} \\
 & SR$\uparrow$ & Steps$\downarrow$ & SR$\uparrow$ & Steps$\downarrow$ & SR$\uparrow$ & Steps$\downarrow$ & SR$\uparrow$ & Steps$\downarrow$ & SR$\uparrow$ & Steps$\downarrow$ \\
\midrule
ACT & \bC{5}{0.960} & \bC{5}{99.0} & \bC{45}{\textbf{0.413}} & \bC{5}{181.3} & \bC{25}{\underline{0.047}} & \bC{25}{\underline{534.7}} & \bC{5}{0.007} & \bC{5}{418.0} & \bC{5}{0.087} & \bC{25}{\underline{455.8}} \\
CoA-r & \bC{45}{\textbf{1.000}} & \bC{25}{\underline{71.6}} & \bC{5}{0.333} & \bC{45}{\textbf{123.3}} & \bC{5}{0.000} & \bC{5}{540.0} & \bC{45}{\textbf{0.013}} & \bC{45}{\textbf{406.9}} & \bC{25}{\underline{0.093}} & \bC{5}{460.4} \\
SkiP & \bC{45}{\textbf{1.000}} & \bC{45}{\textbf{29.8}} & \bC{45}{\textbf{0.413}} & \bC{25}{\underline{169.6}} & \bC{45}{\textbf{0.267}} & \bC{45}{\textbf{435.8}} & \bC{45}{\textbf{0.013}} & \bC{25}{\underline{416.0}} & \bC{45}{\textbf{0.267}} & \bC{45}{\textbf{370.3}} \\
\bottomrule
\end{tabular}\tabularnewline[0.6ex]
\begin{tabular}{l*{5}{C{1.05cm}C{1.15cm}}}
\toprule
\multirow{2}{*}{Method} & \multicolumn{2}{C{2.20cm}}{\TaskUS{put\_money\_in\_safe}} & \multicolumn{2}{C{2.20cm}}{\TaskUS{put\_plate\_in\_dish\_rack}} & \multicolumn{2}{C{2.20cm}}{\TaskUS{reach\_and\_drag}} & \multicolumn{2}{C{2.20cm}}{\TaskUS{screw\_nail}} & \multicolumn{2}{C{2.20cm}}{\TaskUS{setup\_checkers}} \\
 & SR$\uparrow$ & Steps$\downarrow$ & SR$\uparrow$ & Steps$\downarrow$ & SR$\uparrow$ & Steps$\downarrow$ & SR$\uparrow$ & Steps$\downarrow$ & SR$\uparrow$ & Steps$\downarrow$ \\
\midrule
ACT & \bC{45}{\textbf{0.673}} & \bC{25}{\underline{215.6}} & \bC{5}{0.153} & \bC{5}{461.0} & \bC{25}{\underline{0.740}} & \bC{25}{\underline{198.3}} & \bC{5}{0.000} & \bC{5}{450.0} & \bC{5}{0.000} & \bC{5}{640.0} \\
CoA-r & \bC{5}{0.080} & \bC{5}{253.9} & \bC{25}{\underline{0.187}} & \bC{25}{\underline{400.9}} & \bC{5}{0.013} & \bC{5}{329.4} & \bC{45}{\textbf{0.093}} & \bC{45}{\textbf{419.7}} & \bC{5}{0.000} & \bC{5}{640.0} \\
SkiP & \bC{25}{\underline{0.587}} & \bC{45}{\textbf{163.6}} & \bC{45}{\textbf{0.480}} & \bC{45}{\textbf{289.4}} & \bC{45}{\textbf{0.827}} & \bC{45}{\textbf{128.7}} & \bC{25}{\underline{0.053}} & \bC{25}{\underline{423.3}} & \bC{45}{\textbf{0.040}} & \bC{45}{\textbf{595.4}} \\
\bottomrule
\end{tabular}\tabularnewline[0.6ex]
\begin{tabular}{l*{5}{C{1.05cm}C{1.15cm}}}
\toprule
\multirow{2}{*}{Method} & \multicolumn{2}{C{2.20cm}}{\TaskUS{slide\_block\_to\_target}} & \multicolumn{2}{C{2.20cm}}{\TaskUS{straighten\_rope}} & \multicolumn{2}{C{2.20cm}}{\TaskUS{take\_frame\_off\_hanger}} & \multicolumn{2}{C{2.20cm}}{\TaskUS{take\_money\_out\_safe}} & \multicolumn{2}{C{2.20cm}}{\TaskUS{take\_off\_weighing\_scales}} \\
 & SR$\uparrow$ & Steps$\downarrow$ & SR$\uparrow$ & Steps$\downarrow$ & SR$\uparrow$ & Steps$\downarrow$ & SR$\uparrow$ & Steps$\downarrow$ & SR$\uparrow$ & Steps$\downarrow$ \\
\midrule
ACT & \bC{5}{0.307} & \bC{5}{147.4} & \bC{5}{0.000} & \bC{5}{---} & \bC{5}{0.547} & \bC{5}{211.9} & \bC{5}{0.840} & \bC{5}{189.8} & \bC{5}{0.027} & \bC{5}{365.0} \\
CoA-r & \bC{25}{\underline{0.547}} & \bC{25}{\underline{120.1}} & \bC{5}{0.000} & \bC{5}{---} & \bC{45}{\textbf{0.707}} & \bC{45}{\textbf{164.5}} & \bC{45}{\textbf{0.867}} & \bC{25}{\underline{134.0}} & \bC{45}{\textbf{0.093}} & \bC{45}{\textbf{343.8}} \\
SkiP & \bC{45}{\textbf{0.560}} & \bC{45}{\textbf{96.8}} & \bC{5}{0.000} & \bC{5}{---} & \bC{25}{\underline{0.600}} & \bC{25}{\underline{176.5}} & \bC{5}{0.840} & \bC{45}{\textbf{99.2}} & \bC{25}{\underline{0.040}} & \bC{25}{\underline{361.4}} \\
\bottomrule
\end{tabular}\tabularnewline[0.6ex]
\begin{tabular}{l*{5}{C{1.05cm}C{1.15cm}}}
\toprule
\multirow{2}{*}{Method} & \multicolumn{2}{C{2.20cm}}{\TaskUS{take\_plate\_off\_dish\_rack}} & \multicolumn{2}{C{2.20cm}}{\TaskUS{take\_shoes\_out\_of\_box}} & \multicolumn{2}{C{2.20cm}}{\TaskUS{take\_toilet\_roll\_off\_stand}} & \multicolumn{2}{C{2.20cm}}{\TaskUS{take\_umbrella\_out}} & \multicolumn{2}{C{2.20cm}}{\TaskUS{take\_usb\_out\_of\_computer}} \\
 & SR$\uparrow$ & Steps$\downarrow$ & SR$\uparrow$ & Steps$\downarrow$ & SR$\uparrow$ & Steps$\downarrow$ & SR$\uparrow$ & Steps$\downarrow$ & SR$\uparrow$ & Steps$\downarrow$ \\
\midrule
ACT & \bC{25}{\underline{0.733}} & \bC{25}{\underline{206.4}} & \bC{45}{\textbf{0.273}} & \bC{45}{\textbf{582.9}} & \bC{5}{0.533} & \bC{5}{218.5} & \bC{5}{0.360} & \bC{5}{152.8} & \bC{5}{0.880} & \bC{5}{60.2} \\
CoA-r & \bC{5}{0.040} & \bC{5}{328.7} & \bC{25}{\underline{0.013}} & \bC{25}{\underline{653.7}} & \bC{25}{\underline{0.667}} & \bC{25}{\underline{187.1}} & \bC{45}{\textbf{0.560}} & \bC{25}{\underline{125.6}} & \bC{45}{\textbf{0.960}} & \bC{25}{\underline{55.9}} \\
SkiP & \bC{45}{\textbf{0.933}} & \bC{45}{\textbf{89.2}} & \bC{5}{0.000} & \bC{5}{660.0} & \bC{45}{\textbf{0.800}} & \bC{45}{\textbf{111.2}} & \bC{25}{\underline{0.507}} & \bC{45}{\textbf{119.6}} & \bC{25}{\underline{0.933}} & \bC{45}{\textbf{29.5}} \\
\bottomrule
\end{tabular}\tabularnewline[0.6ex]
\begin{tabular}{l*{5}{C{1.05cm}C{1.15cm}}}
\toprule
\multirow{2}{*}{Method} & \multicolumn{2}{C{2.20cm}}{\TaskUS{toilet\_seat\_down}} & \multicolumn{2}{C{2.20cm}}{\TaskUS{toilet\_seat\_up}} & \multicolumn{2}{C{2.20cm}}{\TaskUS{turn\_oven\_on}} & \multicolumn{2}{C{2.20cm}}{\TaskUS{unplug\_charger}} & \multicolumn{2}{C{2.20cm}}{\TaskUS{water\_plants}} \\
 & SR$\uparrow$ & Steps$\downarrow$ & SR$\uparrow$ & Steps$\downarrow$ & SR$\uparrow$ & Steps$\downarrow$ & SR$\uparrow$ & Steps$\downarrow$ & SR$\uparrow$ & Steps$\downarrow$ \\
\midrule
ACT & \bC{5}{0.893} & \bC{5}{102.9} & \bC{25}{\underline{0.840}} & \bC{5}{204.8} & \bC{45}{\textbf{0.647}} & \bC{5}{192.0} & \bC{5}{0.440} & \bC{5}{143.1} & \bC{5}{0.387} & \bC{5}{168.4} \\
CoA-r & \bC{25}{\underline{0.933}} & \bC{25}{\underline{61.8}} & \bC{5}{0.680} & \bC{25}{\underline{177.0}} & \bC{25}{\underline{0.533}} & \bC{25}{\underline{177.8}} & \bC{45}{\textbf{0.720}} & \bC{25}{\underline{109.6}} & \bC{25}{\underline{0.480}} & \bC{25}{\underline{142.3}} \\
SkiP & \bC{45}{\textbf{1.000}} & \bC{45}{\textbf{40.8}} & \bC{45}{\textbf{0.880}} & \bC{45}{\textbf{99.5}} & \bC{5}{0.453} & \bC{45}{\textbf{175.4}} & \bC{25}{\underline{0.640}} & \bC{45}{\textbf{107.1}} & \bC{45}{\textbf{0.533}} & \bC{45}{\textbf{120.5}} \\
\bottomrule
\end{tabular}
\end{longtable}
\endgroup

\section{Limitations and Discussion}
\label{sec:appendix:limitations}
\label{sec:limitations}

\label{sec:exp:failure}

\textbf{Failure modes.}
We identify three cases where SkiP underperforms: (1)~\emph{hard constraint tasks} (\TaskUS{beat\_the\_buzz}), where instant-fail contact conditions penalize even small positional overshoots during skip jumps; (2)~\emph{over-fragmented segmentation} (\TaskUS{take\_shoes\_out\_of\_box}), where long repetitive pick-and-place sequences produce too many short segments; (3)~\emph{precision continuous manipulation} (\TaskUS{turn\_oven\_on}), where smooth rotational motions still require step-level precision. These cases share a common pattern: low DCT frequency does not always mean the trajectory is safe to skip. Detailed per-task analysis is in Appendix~\ref{sec:appendix:failure}.

\textbf{Label source dependence.} All main results use MSK for segmentation. We compare against alternative label sources in \S\ref{sec:exp:ablation} (Table~\ref{tab:label_source}), but have not tested oracle contact labels or learned segmenters. The gap between MSK and the best alternative (LV, $+0.078$) suggests room for better segmentation to further improve SkiP.

\section{Action Displacement Analysis}
\label{sec:appendix:action_disp}

For each policy call during evaluation, we measure the \emph{jump distance}: $\|a_{1} - p_{\text{ee}}\|_2$, the Euclidean distance between the first predicted target position and the current end-effector position.
If SkiP learns distinct skip and refine modes, we expect a bimodal distribution: large jumps when the policy skips past free space, and small adjustments when it refines near contacts.

Figure~\ref{fig:action_disp} (main paper) shows this distribution across $10$ RLBench tasks.
For SkiP, we split policy calls into \emph{key} and \emph{skip} categories using a per-task displacement threshold derived from training-demo statistics.
SkiP exhibits a clear bimodal pattern: skip-mode calls produce jumps of $0.1$--$0.7$\,m, while key-mode calls cluster near zero.
In contrast, CoA-rev (which uses temporal ensembling with per-step replanning) concentrates near zero across all tasks, and CoA-fwd (open-loop long chunks) shows moderate but unimodal displacements.
This supports the interpretation that relabeling teaches the policy to produce qualitatively different outputs depending on the trajectory phase, rather than a single averaged behavior.

\section{Supplementary Ablation Details}
\label{sec:appendix:ablation_details}

\paragraph{Key-segment component contribution.}
The $0.076$-point SR gap between the full MSK variant ($0.606$) and DCT-only ($0.530$) on RLBench-60 quantifies how much of SkiP's performance is attributable to the specific labeling stack (heuristic keyframe union and bend-based augmentation).
Each component contributes independently, and the full variant achieves the best SR.
Notably, even the DCT-only variant ($0.530$) is comfortably above CoA-rev's $0.490$ on the same suite, confirming that the relabeling mechanism itself, not the labeling refinements, accounts for the majority of the gain.

\paragraph{Cross-representation transfer.}
MSK transfers across action representations: on RLBench-60, joint-space DCT achieves $0.568$ SR versus $0.574$ for end-effector DCT (both without bend, since bend requires end-effector coordinates).
The negligible gap ($0.006$) indicates that frequency content carries similar task-relevant information regardless of the action space, and practitioners can apply MSK to whichever representation is available.

\paragraph{Label source analysis.}
The poor performance of Velocity Only (VO) on sustained-contact tasks (\Task{open-box}: $0.027$, \Task{sweep-dustpan}: $0.040$) occurs because high-velocity timesteps rarely coincide with manipulation-critical phases; these tasks involve slow, careful motions that VO mislabels as skippable.
Low Velocity Key (LV, $0.773$) is the strongest alternative, consistent with the intuition that slow motion correlates with careful manipulation, but it still misses trajectory curvature and frequency-domain patterns that MSK captures through spectral decomposition.
Random Stride (RS, $0.520$) performs poorly despite matching the ${\sim}25\%$ key-segment ratio, proving that segment \emph{location} matters more than segment \emph{count}.

\paragraph{Absolute-target requirement (expanded).}
SkiP assumes the controller can reach a distant waypoint in one step, which holds when actions are absolute poses or joint angles.
All four benchmarks satisfy this: motion planning on RLBench, absolute end-effector pose targets on RoboMimic, absolute joint targets on RoboTwin and $\pi_{0.5}$.
Under delta-action or velocity control, skip targets would need to be reformulated as accumulated deltas over the skipped span, and large gaps may introduce tracking errors that offset the benefit of skipping.
Extending SkiP to delta-action settings is a natural direction for future work.

\section{Failure Case Study Details}
\label{sec:appendix:failure}

We expand on the three failure modes identified in \S\ref{sec:exp:failure} with quantitative segment statistics and root-cause analysis.

\paragraph{Setup.}
All evaluations use the main SkiP configuration (W=16, q=0.75, $L_\text{min}$=3). Segment statistics are computed over 20 training demonstrations per task. Each task is evaluated with 3 repeats $\times$ 25 episodes.

\begin{table}[h]
\centering
\caption{DCT segment statistics for three failure tasks vs.\ two representative success tasks.}
\label{tab:failure_segments}
\small
\begin{tabular}{lccccc}
\toprule
Task & Ep.\ Length & Segments & Avg Seg Len & Key Ratio \\
\midrule
\TaskUS{beat\_the\_buzz} & 138.1 & 4.9 & 7.5 & 26.7\% \\
\TaskUS{take\_shoes\_out\_of\_box} & 465.2 & 20.0 & 6.2 & 26.6\% \\
\TaskUS{turn\_oven\_on} & 132.4 & 4.3 & 10.1 & 31.0\% \\
\midrule
\TaskUS{open\_door} (success) & $\sim$90 & 3--4 & $\sim$25 & $\sim$30\% \\
\TaskUS{lamp\_off} (success) & $\sim$70 & 2--3 & $\sim$30 & $\sim$28\% \\
\bottomrule
\end{tabular}
\end{table}

\paragraph{(1) Hard constraint violation: \TaskUS{beat\_the\_buzz}.}
The task requires sliding a wand along a curved wire from left to right; touching the wire triggers immediate episode termination. The trajectory is geometrically smooth (low DCT spectral energy), so MSK labels most of the wire-following phase as a skip segment. However, the wand must maintain sub-centimeter clearance from the wire at every step. When SkiP issues a skip action that jumps several steps ahead, the resulting straight-line interpolation can clip the wire, terminating the episode. CoA-rev, which replans at every step, maintains the fine-grained corrections needed to avoid contact (SR 0.467 vs.\ SkiP 0.320).

\paragraph{(2) Over-fragmented segmentation: \TaskUS{take\_shoes\_out\_of\_box}.}
This task requires extracting two shoes from a box via 4+ pick-and-place cycles, producing the longest episodes in RLBench-60 (465 steps). The repetitive grasp-lift-move-release pattern creates alternating frequency signatures, yielding 20 segments averaging only 6.2 steps each. This extreme fragmentation causes: (a) compounding errors at each skip-attend transition, and (b) phase confusion where the policy cannot track which sub-goal is current. All 75 evaluation episodes time out at the 660-step budget. Notably, ACT (simple action chunking without skip/CoA decomposition) achieves 27.3\% SR, suggesting that long multi-object tasks benefit from uniform temporal resolution.

\paragraph{(3) Precision continuous manipulation: \TaskUS{turn\_oven\_on}.}
Rotating a knob through 86$^{\circ}$ (1.5 rad) requires the end-effector to maintain contact and apply consistent rotational force. The segmentation is well-behaved (4.3 segments, 10.1 avg length), so failure is not due to fragmentation. Rather, the rotation phase is smooth and repetitive (low frequency), causing MSK to label it as skippable. Skipping within this phase causes loss of contact or angular overshoot. When SkiP fails, episodes consistently time out at 280 steps, indicating the policy gets close to but cannot precisely reach the 1.5 rad threshold.

\paragraph{Common thread and future directions.}
All three cases share a root cause: the frequency heuristic assumes that smooth (low-frequency) motion is safe to skip, but this assumption fails when:
(a)~smooth trajectories have hard safety constraints (no-contact zones),
(b)~repetitive motion patterns fragment the DCT into many short segments, or
(c)~smooth motion requires sustained precision to meet an exact threshold.

Potential mitigations include: (1) \emph{constraint-aware skipping} that incorporates task-level safety boundaries into the skip decision; (2) \emph{adaptive segment merging} that detects and consolidates over-fragmented patterns in repetitive tasks; and (3) \emph{learned skip confidence} that attenuates skip magnitude during high-precision manipulation phases.

\end{document}